\documentclass[11pt]{article}
\usepackage[preprint]{acl}

\usepackage{times}
\usepackage{latexsym}

\usepackage[T1]{fontenc}

\usepackage[utf8]{inputenc}

\usepackage{microtype}

\usepackage{inconsolata}

\usepackage{graphicx}

\usepackage{arydshln} 
\usepackage[whole]{bxcjkjatype} 
\usepackage{multicol}
\usepackage{multirow}
\usepackage{booktabs}
\usepackage{url}
\usepackage{hyperref}
\usepackage[most]{tcolorbox}
\usepackage[table]{xcolor}
\usepackage{bbm}
\usepackage{amsmath}
\usepackage{amssymb}
\usepackage{framed}
\title{Uncovering Superficial Cues That Affect Confidence in LLM Reasoning}

\newcommand{\red}[1]{\textcolor{black}{#1}}

\author{
  \begin{tabular}{ccc}
    Shintaro Ozaki$^{1}$ &
    Kazuki Hayashi$^{1}$ &
    Wataru Hashimoto$^{1}$ \\[0.8em]
    Hidetaka Kamigaito$^{1}$ &
    Katsuhiko Hayashi$^{2,1}$ &
    Taro Watanabe$^{1}$
  \end{tabular}
  \\[1.5em]
  $^{1}$ Nara Institute of Science and Technology (NAIST), Japan \\
  $^{2}$ The University of Tokyo, Japan
  \\[0.3em]
  \texttt{\{ozaki.shintaro.ou6, kamigaito.h, taro.watanabe\}@naist.ac.jp} \\
}

\begin{document}
\maketitle
\begin{abstract}
While explicit reasoning improves the performance of large language models (LLMs), their responses are often overconfident, even though they include linguistic expressions demonstrating uncertainty.
Many studies have attempted to mitigate model overconfidence, whereas it has not been investigated what linguistic expressions influence confidence in reasoning.
To address this gap, our work identifies what linguistic expressions are related to confidence.
Specifically, we apply regression analysis with confidence as the dependent variable to quantify the relationship between individual $n$-grams and model confidence.
Across multiple models and QA benchmarks, we show that LLMs remain overconfident when reasoning is involved and attribute this behavior to specific linguistic expressions.
Interestingly, several of the extracted expressions coincide with cue phrases intentionally inserted during test-time scaling to improve reasoning performance.
These tendencies are consistent across multiple analyses, indicating that the $n$-grams do not depend on the downstream task.
Moreover, we show that the $n$-grams affecting confidence do not always align with human intuition, highlighting the importance of identifying confidence-related $n$-grams.
\end{abstract}

\begin{figure*}[t]
    \centering
    \includegraphics[width=\textwidth]{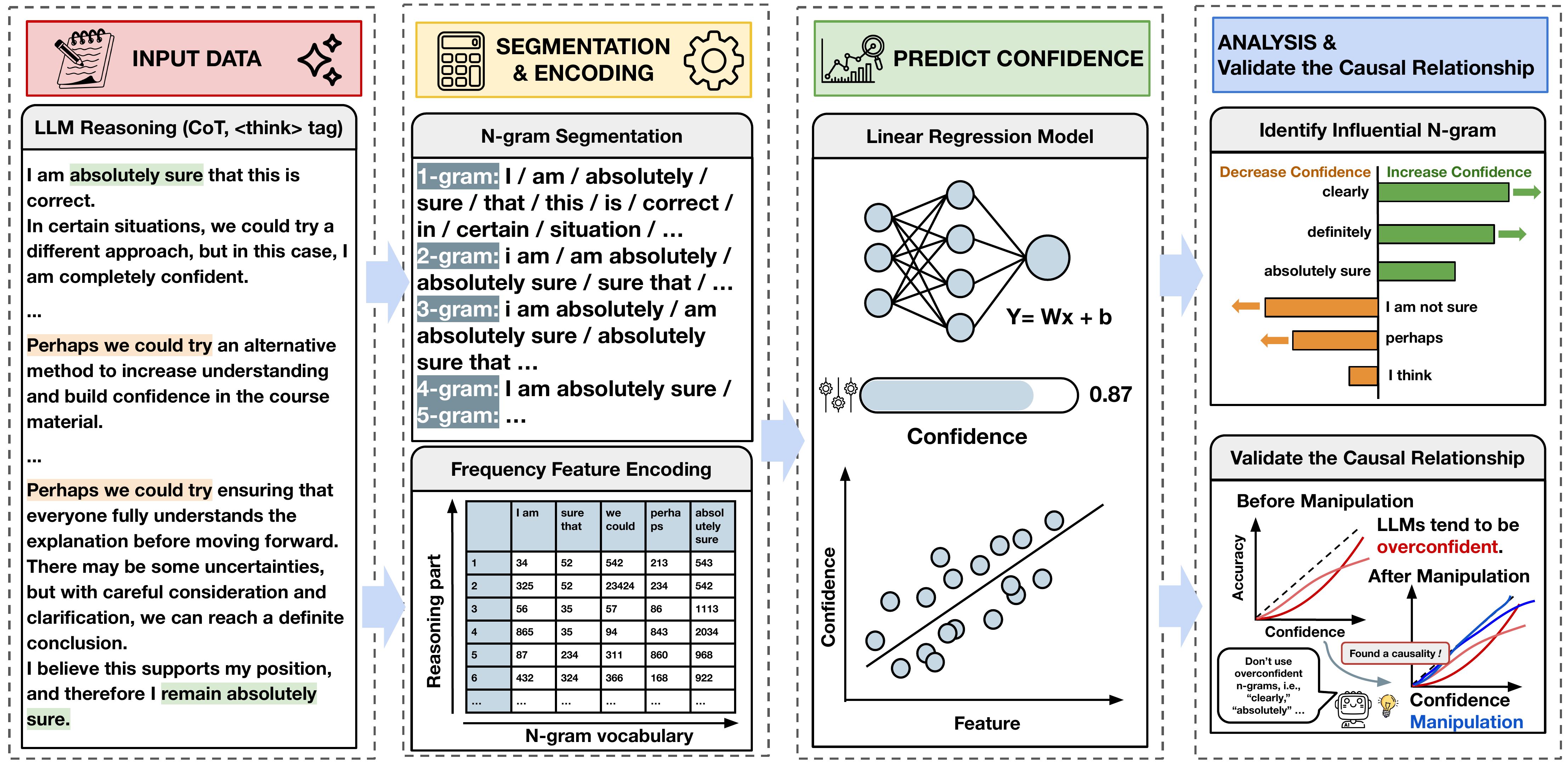}
    \caption{
    An overview of our work.
    We segment reasoning processes into $n$-grams to identify those contributing to overconfidence, and manipulate internal reasoning via prompting to verify their causal effect on confidence.
    }
    \label{fig:top}
\end{figure*}

\section{Introduction}
While explicit reasoning, such as Chain-of-Thought~\cite[CoT;][]{wei2022chain,kojima2022large,brown2020language} and \texttt{<think>} tag augmented reasoning processes by, e.g., DeepSeek-R1~\citep{guo2025deepseek}, improve the performance of large language models (LLMs), LLMs are known to remain overconfident; i.e., they produce final answers with high confidence despite the answers being incorrect.
In particular, even when reasoning processes contain uncertain linguistic expressions such as ``it is difficult to decide between A or B'' or ``the information is insufficient,'' models often show overconfidence~\citep{kirichenko2025abstentionbench, xiong2024can, xia-etal-2025-survey}.
This leads to hallucinations~\citep{xiong2024can,ji-etal-2025-calibrating,liu2025uncertainty}, which present a critical problem for AI/ML safety~\citep{amodei2016concrete, hendrycks2021unsolved}.
Many studies attempt to identify \red{where this confidence comes from} in terms of the model output~\cite{wang2023selfconsistency, zhou2023leasttomost, mielke-etal-2022-reducing}, whereas how linguistic expressions appearing within the reasoning part affect the formation of model confidence has remained unrevealed~\citep{stolfo2024confidence, ji-etal-2025-calibrating, tian-etal-2023-just}.
As a result, they overlook the seemingly superficial linguistic cues in shaping model confidence.%
\footnote{We define \textit{superficial cues} as $n$-grams that appear to have little or no effect on model confidence from a human perspective, but in fact have a large influence on the confidence.}
In short, they leave the research question: ``\textit{why models become overconfident or what linguistic expressions shape confidence.}''

To reveal the superficial cues, we focus on CoT processes in instruction-tuned models~\citep{wei2022chain,kojima2022large,brown2020language} as well as reasoning processes enclosed in \texttt{<think>} tags in reasoning models~\citep{guo2025deepseek, qwen_qwq32b_2025} to analyze the relationship between linguistic features and model confidence.
Specifically, as shown in the left part of Figure~\ref{fig:top}, we split the generated reasoning processes into $n$-grams and perform regression analysis~\citep{tibshirani1996regression} with confidence as the dependent variable.
This clarifies what $n$-grams affect underconfidence, i.e., expressing low confidence in the final output despite the answer being correct, or overconfidence.

Experiments using multiple models on multiple QA tasks show that generating specific $n$-grams tends to excessively increase confidence and cause models to act overconfidently.
We then find that specific linguistic expressions for test-time scaling, i.e., extending reasoning time to gain performance by intentionally inserting ``Wait,'' or ``Hmm,''  also affect confidence.
These tendencies are consistent across multiple analysis methods, i.e., forced decoding and generation-based approaches, and task formats, i.e., multiple-choice and open-ended QA, suggesting that the identified $n$-grams generalize across both methods and downstream tasks.
We also show that the $n$-grams affecting confidence do not always align with human intuition about uncertainty~\cite{clark2002using}, e.g., ``Uh,'' ``I am unsure'', highlighting the importance of identifying the confidence-related $n$-grams.

\section{Related Work}
Research investigating confidence has remained a vital perspective for the LLMs era~\citep{xiong2024can,pal-etal-2023-med,10.1145/3703155,zhang2025siren}.
In fields where information transparency and reliability are critical, such as finance~\citep{xu2025profit}, medicine~\citep{lambert2024trustworthy}, and autonomous driving~\citep{e26080634,hoel2023ensemble}, one must account for confidence~\citep{geng-etal-2024-survey, zhang2025survey}. 

LLMs often show overconfidence in their outputs~\citep{xiong2024can,pal-etal-2023-med,10.1145/3703155,zhang2025siren}.
Prior works identify instruction-tuning~\citep{zhang2025instructiontuninglargelanguage} and RLHF~\citep{christiano2017deep} as primary causes~\citep{achiam2023gpt,tian-etal-2023-just,kadavath2022language,hu2025navigatingalignmentcalibrationtradeoffparetosuperior}.
While these processes modify the output distribution to favor human preferences, safety, and instruction-following, which often prevents the output probabilities from reflecting the true probability of correctness~\citep{achiam2023gpt,zhou2025steerconf}, they have not focused on what specific linguistic expressions affect confidence.
\citet{vanhoyweghen2025lexicalhintsaccuracyllm} further showed that lexical uncertainty markers in CoT correlate with response correctness and can serve as lightweight post-hoc calibration signals.
\red{While their motivation aligns closely with ours in leveraging lexical hints rather than output probabilities to assess accuracy, their work does not explicitly focus on confidence.}

Many studies aim to mitigate model overconfidence by modifying confidence estimates after or during inference.
For example, \citet{stolfo2024confidence} and \citet{ji-etal-2025-calibrating} intervene in the internal states of LLMs, while \citet{tian-etal-2023-just} and \citet{guo2017calibration} adjust output probabilities through post-hoc identification techniques.
Other approaches improve confidence reliability by incorporating external evidence, particularly through Retrieval-Augmented Generation (RAG)~\citep{lewis2020retrieval,liu-etal-2025-abstain,pmlr-v239-ren23a,tao-etal-2024-trust}.
However, these approaches primarily focus on modifying confidence estimates rather than identifying what causes confidence to arise.
Consequently, the linguistic factors that lead LLMs to become overconfident or underconfident during reasoning remain largely unexplored.

\section{Methods}
\label{methods}
To determine what linguistic expressions affect output confidence, we calculate confidence scores (\S\ref{confidence}) and perform a regression analysis using these scores as the dependent variable and the reasoning process as the features (\S\ref{regression}).
Then, we verify the causal relationship by manipulating the LLMs using the linguistic expressions identified through the regression analysis (\S\ref{re-generation}).

\subsection{Confidence Calculation}
\label{confidence}
Regarding the calculation of confidence, two primary methods exist: (1) the generation-based method~\citep{xiong2024can,mielke2022reducing} and (2) the forced-decoding method~\citep{kuhn2023semantic,duan-etal-2024-shifting,mielke2022reducing}.
In our work, we primarily calculate confidence using the forced-decoding method because evaluation results for generation-based methods vary across studies~\citep{chen2023meditron,wu2023pmc}.%
\footnote{We also calculate the confidence by the generation-based method, and observe the same tendency in Appendix~\ref{calcurating-confidence-by-generation-or-force-decoding}.}
Specifically, we append the phrase ``\texttt{So, the answer is}'' to the reasoning part, following previous studies~\citep{ozaki-etal-2025-understanding,wang-etal-2025-chain}, and compute the probability of each option:
\begin{align}
    P(x_i) &= \frac{P(x_i \mid \text{prompt})}
    {\sum_{j=1}^{J} P(x_j \mid \text{prompt})}.
    \label{eq:confidence}
\end{align}
Here, $x_i$ represents the token corresponding to the $i$-th option, and $J$ is the total number of options.
$P(x_i \mid \text{prompt})$ denotes the model probability of generating option $x_i$ given the prompt, which includes the reasoning part and ``\texttt{So, the answer is}.'' 
The probabilities are normalized across all options to obtain $P(x_i)$.
Our work uses the confidence score of the final option selected by the model, i.e., $P(x_i)$ for the generated answer.

\subsection{Influential $N$-gram Estimation in Confidence and Accuracy}
\label{regression}
To quantitatively analyze the relationship between the linguistic features in the reasoning process and the model's predictive confidence, we apply a regression analysis based on $n$-grams, using the scikit-learn library~\citep{scikit-learn}. 
Specifically, each data point consists of a reasoning process generated by the model and the output probability for the answer selected based on that reasoning. 
We define the confidence score $y_i$ as the probability corresponding to the model's final predicted option, treated as a continuous value in the range $[0, 1]$, as introduced in \S\ref{confidence}.
As preprocessing, we lowercase the reasoning segments and extract tokens consisting only of alphabetic characters. 
Subsequently, we create features using the frequency of $n$-grams (from $n=1$ to $5$, as we follow the conventional $n$-gram approach) that appear at least twice.%
\footnote{Appendix~\ref{app:sensitivity-to-feature-representation} shows the comparison with TF-IDF or binary-based approach. Our approach showed a high correlation.}
To avoid frequency bias, i.e., 1-grams are more frequent than 5-grams, we train a different regression model for each $n$-gram size and finally compare the coefficient value associated with each $n$-gram using Z-score standardization.
Further details, such as cross-validation and loss results, are in the Appendix~\ref{detailed-experimental-setings}.

We convert each reasoning segment from the dataset consisting of $D$ reasoning segments into a feature vector $\mathbf{f}_i \in \mathbb{R}^d$ derived from $n$-gram frequencies, and let $\mathbf{F} \in \mathbb{R}^{D \times d}$ be the matrix where these vectors serve as rows. 
A linear regression model $\hat{y}_i = \mathbf{f}_i^\top \boldsymbol{\beta}$ represents the relationship between confidence and $n$-gram features, where $\boldsymbol{\beta} \in \mathbb{R}^d$ is the vector of regression coefficients corresponding to each $n$-gram. 
We apply L2 normalization to $\mathbf{f}_i$ so that the sentence length of the reasoning part does not bias the weights, i.e., $\tilde{\mathbf f}_i = \mathbf f_i / \|\mathbf f_i\|_2$.
Following \citet{chahuneau-etal-2012-word}, we employ Lasso regression~\citep{tibshirani1996regression} to extract features:
\begin{align}
\hat{\boldsymbol{\beta}} &= \arg\min_{\boldsymbol{\beta}}\left(\frac{1}{2D}\sum_{i=1}^{D}\left(y_i - \tilde{\mathbf f}_i^\top \boldsymbol{\beta}\right)^2 + \lambda \|\boldsymbol{\beta}\|_1\right). 
\end{align}
Here, $\lambda$ represents the regularization coefficient and $\lVert \boldsymbol{\beta} \rVert_1$ is the L1 norm.

Based on the sign and magnitude of the estimated regression coefficients, we extract $n$-grams that lead to increasing or decreasing confidence, which quantifies the linguistic expressions that influence the model's tendency toward high or low confidence, respectively.

Furthermore, by using logistic regression~\citep{cox1958regression} instead of Lasso regression and treating a binary label prediction accuracy as the dependent variable, we identify the $n$-grams that influence accuracy.
Further details are provided in Appendix~\ref{detailed-experimental-setings}.

\begin{figure}[t]
    \centering
    \includegraphics[width=\linewidth]{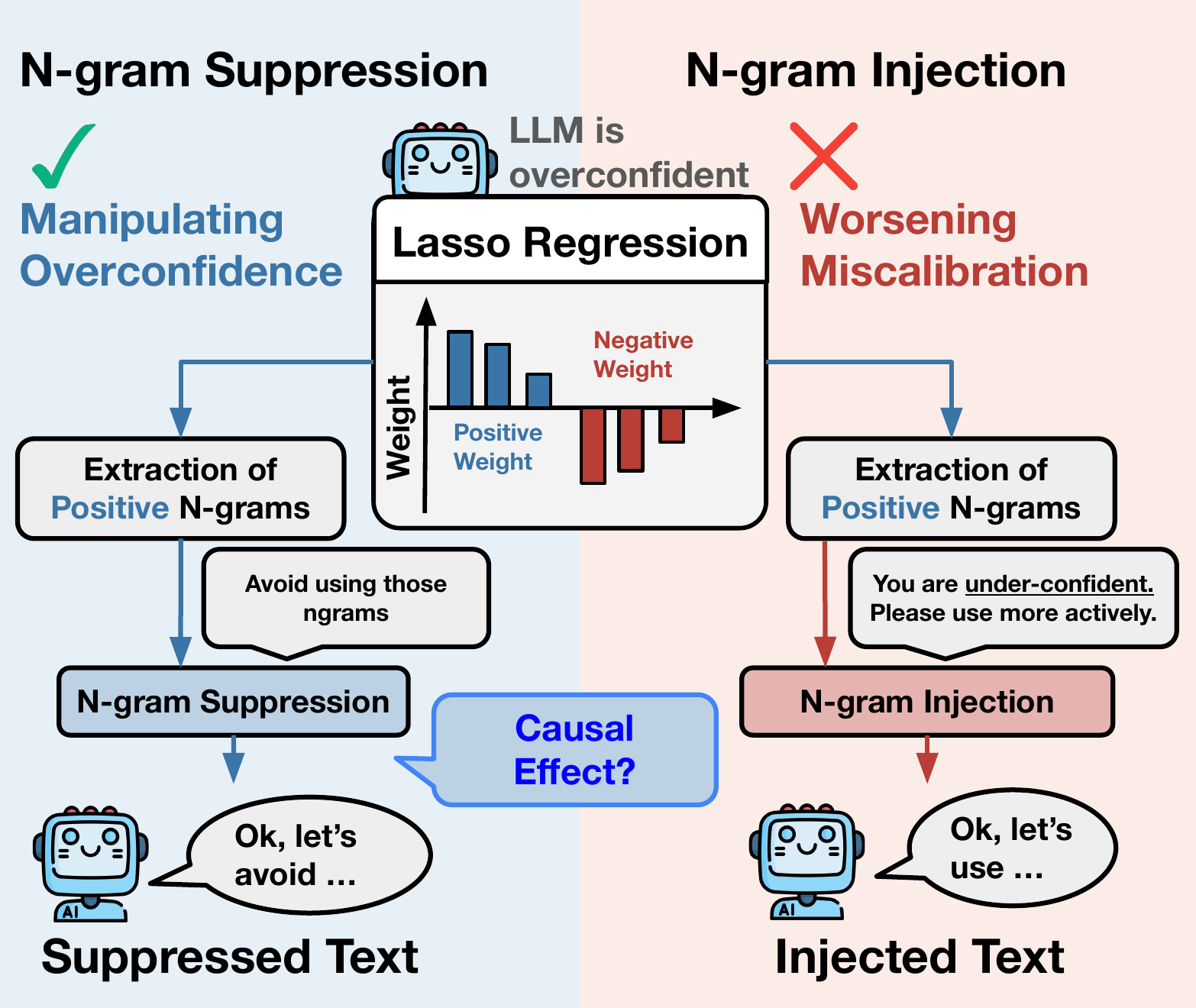}
    \caption{To confirm the causal effect, i.e., the $n$-grams should affect the confidence, we try manipulating them.}
    \label{fig:calibration-approach}
\end{figure}

\subsection{Re-generation with Influential $N$-gram}
\label{re-generation}
While the identified $n$-grams may indicate overconfidence, the cause of overconfidence is not necessarily limited to $n$-grams, i.e., factors other than $n$-grams may affect overconfidence.

To reveal that the identified $n$-grams truly affect confidence, we examine whether model confidence can be manipulated through two complementary interventions based on the linguistic expressions identified by Lasso regression, thereby assessing their causal influence on confidence.

Specifically, since previous work has shown that LLMs tend to be overconfident~\citep{kirichenko2025abstentionbench,xiong2024can,pal-etal-2023-med,10.1145/3703155,zhang2025siren,wang-etal-2025-language-mixing}, we first investigate whether confidence can be reduced by preventing the model from using linguistic expressions that Lasso identifies as increasing confidence, as illustrated in Figure~\ref{fig:calibration-approach}.
We extract the top-$k$ $n$-grams with positive coefficients and instruct the LLMs to avoid generating them.
We refer to this intervention as \underline{$N$-gram Suppression}, and adopt $k=5$, which achieves the best performance.%
\footnote{
We compare different variations in Appendix~\ref{app:differentiation-by-prompting-methods}, e.g., prompts that do not explicitly mention overconfidence.
}

In addition, we investigate the opposite intervention by encouraging the model to use the same confidence-increasing $n$-grams.
We refer to this intervention as \underline{$N$-gram Injection}.
Appendix~\ref{detailed-prompts} provides all prompts used in our experiments.

\section{Experimental Settings}
\paragraph{Models.}
For the reasoning models, we use 5 reasoning models: DeepSeek-R1 models distilled from Qwen~\citep{guo2025deepseek, hinton2015distilling}, denoted as R1 Qwen, Qwen3, Phi-4 (reasoning)~\citep{abdin2025phi}, GPT-oss~\citep{agarwal2025gpt}, and Nemotron~\citep{bercovich2025llama}.
Furthermore, for the instruction-tuned models that generate reasoning processes through CoT, we use 2 models: Qwen2.5~\citep{qwen2024qwen2}, and Phi-4 (mini)~\citep{abouelenin2025phi}, resulting in a total of 7 models for our experiments.
Appendix~\ref{detailed-experimental-setings} provides the detailed settings and model names.

\paragraph{Datasets.}
Our work utilizes multiple-choice QA, which allows for confidence calculation since each answer option can be represented as a single token with an associated probability.%
\footnote{We conduct the comparison analysis between the multiple-choice and open-end generation QAs in \S\ref{results-and-discussions}.}
Specifically, we use 5 datasets: MathQA~\citep{amini-etal-2019-mathqa}, MMLU~\citep{hendryckstest2021}, HellaSwag~\citep{zellers-etal-2019-hellaswag}, RACE~\citep{lai-etal-2017-race}, and CosmosQA~\citep{huang-etal-2019-cosmos}.
MathQA contains 5 options, while the other datasets contain 4.
The total size is 21,500, and Appendix~\ref{detailed-dataset} provides detailed settings.

\paragraph{Evaluation Metrics.}
\label{evaluation-metrics}

We use Expected Calibration Error (ECE)~\citep{naeini2015obtaining} and Adaptive Calibration Error (ACE)~\citep{nixon2019measuring}, which are widely used and standard metrics for quantitatively evaluating the discrepancy between confidence and accuracy.
To provide a more comprehensive quantitative evaluation, we also include Brier Score~\citep{glenn1950verification} for evaluating the calibration of predicted probabilities, AUC-ROC~\citep{bradley1997use} for evaluating discriminatory performance independent of a threshold, and Accuracy.
Note that higher values are better for Accuracy and AUC-ROC, while lower values are better for the other metrics, i.e., ECE, ACE, and Brier Score.
Appendix~\ref{detailed-evaluation-metrics} provides explanations.

\begin{figure*}[t]
    \centering
    \includegraphics[width=\textwidth]{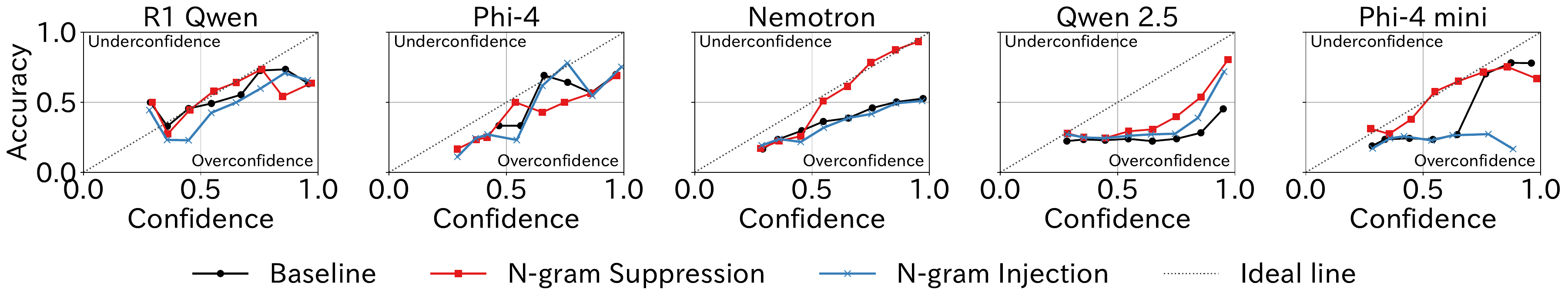}
    \caption{
    Confidence behavior before and after manipulating confidence-related $n$-grams.
    Ideally, the curve matches the diagonal line ($y=x$), meaning that when the x-axis confidence (output probability) is 70\%, the y-axis accuracy should also be 70\%.
    }
    \label{fig:calibration-curve}
\end{figure*}

\begin{table*}[t]
    \centering
    \setlength{\tabcolsep}{3pt}
    \renewcommand{\arraystretch}{1.2}
    \rowcolors{6}{gray!10}{white}
    \resizebox{\linewidth}{!}{
    \begin{tabular}{l ccccc@{\hspace{8pt}} ccccc@{\hspace{8pt}} ccccc}
        \toprule
    \multirow{2.25}{*}{\textbf{Model}} & \multicolumn{5}{c}{\textbf{Baseline Inference}} & \multicolumn{5}{c}{\textbf{$N$-gram Suppression}} & \multicolumn{5}{c}{\textbf{$N$-gram Injection}} \\
    \cmidrule(lr){2-6}\cmidrule(lr){7-11}\cmidrule(lr){12-16}
     & \textbf{ECE $\downarrow$} & \textbf{ACE $\downarrow$} & \textbf{Brier $\downarrow$} & \textbf{Acc. $\uparrow$} & \textbf{AUROC $\uparrow$}
     & \textbf{ECE $\downarrow$} & \textbf{ACE $\downarrow$} & \textbf{Brier $\downarrow$} & \textbf{Acc. $\uparrow$} & \textbf{AUROC $\uparrow$}
     & \textbf{ECE $\downarrow$} & \textbf{ACE $\downarrow$} & \textbf{Brier $\downarrow$} & \textbf{Acc. $\uparrow$} & \textbf{AUROC $\uparrow$} \\
    \midrule
\rowcolor{blue!10}
\multicolumn{16}{l}{\textbf{Reasoning Models}} \\
R1 Qwen & 0.334 & 0.334 & 0.713 & 0.627 & 0.502 & 0.318 & 0.316 & \textbf{0.700\textsuperscript{*}} & \textbf{0.630\textsuperscript{*}} & 0.510 & \textbf{0.312} & \textbf{0.310} & 0.704 & 0.616 & \textbf{0.542} \\
R1 Qwen3 & 0.198 & 0.198 & 0.450 & 0.759 & \textbf{0.604} & \textbf{0.166\textsuperscript{*}} & \textbf{0.166\textsuperscript{*}} & \textbf{0.344\textsuperscript{*}} & \textbf{0.826\textsuperscript{*}} & 0.566 & 0.167 & 0.166 & 0.359 & 0.815 & 0.566 \\
Nemotron & 0.415 & 0.415 & 0.896 & 0.510 & 0.517 & \textbf{0.334\textsuperscript{*}} & \textbf{0.334\textsuperscript{**}} & \textbf{0.859\textsuperscript{**}} & \textbf{0.553\textsuperscript{**}} & \textbf{0.637\textsuperscript{**}} & 0.409 & 0.409 & 0.898 & 0.496 & 0.544 \\
Phi-4 & 0.199 & 0.199 & \textbf{0.405} & \textbf{0.795} & 0.562 & \textbf{0.194} & \textbf{0.194} & 0.448 & 0.734 & \textbf{0.689\textsuperscript{*}} & 0.209 & 0.209 & 0.486 & 0.714 & 0.672 \\
GPT-oss & 0.191 & 0.191 & 0.412 & 0.783 & 0.532 & \textbf{0.191} & \textbf{0.191} & \textbf{0.401\textsuperscript{**}} & \textbf{0.795\textsuperscript{**}} & \textbf{0.546\textsuperscript{**}} & 0.197 & 0.197 & 0.412 & 0.791 & 0.526 \\
\rowcolor{red!10}
\multicolumn{16}{l}{\textbf{Instruction-tuned Models}} \\
Qwen 2.5 & 0.329 & 0.329 & 0.944 & 0.274 & 0.587 & \textbf{0.205\textsuperscript{**}} & \textbf{0.205\textsuperscript{**}} & \textbf{0.657\textsuperscript{**}} & \textbf{0.501\textsuperscript{**}} & \textbf{0.777\textsuperscript{**}} & 0.245 & 0.245 & 0.849 & 0.297 & 0.600 \\
Phi-4 mini & 0.285 & 0.283 & \textbf{0.640} & \textbf{0.660} & 0.509 & \textbf{0.128\textsuperscript{**}} & \textbf{0.127\textsuperscript{**}} & 0.720 & 0.654 & \textbf{0.711\textsuperscript{**}} & 0.132 & 0.132 & 0.809 & 0.245 & 0.510 \\
    \bottomrule
    \end{tabular}
    }
    \caption{Experimental results. Bold text indicates the best score among the models. Superscripts indicate statistically significant improvements over the baseline (*: $p<0.05$, **: $p<0.01$). AUROC is an abbreviation of AUC-ROC.
    We perform bootstrap resampling~\cite{koehn-2004-statistical} with 1,000 resamples.
    }
    \label{tab:overall-results}
\end{table*}

\begin{table}[t]
    \centering
    \setlength{\tabcolsep}{2pt}
    \resizebox{\linewidth}{!}{
    \begin{tabular}{lccccc}
    \toprule
    \textbf{Metric} & \textbf{1-gram} & \textbf{2-gram} & \textbf{3-gram} & \textbf{4-gram} & \textbf{5-gram} \\
    \midrule
    Pearson     & 0.61 & 0.68 & 0.53 & 0.59 & 0.34 \\
    Jaccard@50 & 0.85 & 0.81 & 0.80 & 0.73 & 0.49 \\
    Jaccard@100 & 0.72 & 0.69 & 0.64 & 0.46 & 0.40 \\
    \bottomrule
    \end{tabular}
    }
    \caption{Correlation of $n$-grams obtained from generation and forced-decoding methods. @$k$ denotes top-$k$.}
    \label{tab:correlation-bw-generation-and-forceddecoding}
\end{table}

\section{Results and Discussion}
\label{results-and-discussions}
Figure~\ref{fig:calibration-curve} shows the confidence trends, and Table~\ref{tab:overall-results} presents the evaluation results using the metrics described in \S\ref{evaluation-metrics}. 
To interpret Figure~\ref{fig:calibration-curve}, we divide the confidence scores into $n$ bins and calculate the accuracy within each bin.%
\footnote{Bin size is 10 in our work, following previous studies~\citep{ozaki-etal-2025-understanding, guo2017calibration, nixon2019measuring}.}
Ideally, the curve should be identical to the diagonal line, meaning that when, e.g., the x-axis confidence is 70\%, the y-axis accuracy should be 70\%.
The results in Figure~\ref{fig:calibration-curve} show that all models remain overconfident, where accuracy is low despite high predicted probabilities. 
This finding aligns with the results reported in prior work~\citep{xiong2024can,pal-etal-2023-med,10.1145/3703155,zhang2025siren}. 
In addition, Table~\ref{tab:overall-results} indicates that suppressing confidence-related $n$-grams changes confidence estimates, thus revealing a causal effect.

\begin{table*}[t]
    \centering
    \setlength{\tabcolsep}{2pt}
    \rowcolors{4}{gray!10}{white}
    \resizebox{\textwidth}{!}{
    \begin{tabular}{crrr crrr crrr crrr}
    \toprule
    \textbf{1-gram} &  \textbf{Conf.} & \textbf{Acc.} & \textbf{Freq.} &
    \textbf{2-gram} &  \textbf{Conf.} & \textbf{Acc.} & \textbf{Freq.} &
    \textbf{3-gram} &  \textbf{Conf.} & \textbf{Acc.} & \textbf{Freq.} &
    \textbf{4-gram} &  \textbf{Conf.} & \textbf{Acc.} & \textbf{Freq.} \\
\cmidrule(lr){1-16}
\rowcolor{blue!10}
\multicolumn{16}{c}{\textbf{Top 10 $n$-grams that boost the confidence.}} \\
\cmidrule(lr){1-16}
looking & 2.16 & 0.00$^\dagger$ & 27776 & options answer & 2.35 & 0.00$^\dagger$ & 972 & options hmm let & 1.81 & 0.14$^\dagger$ & 79 & correct answer wait let & 1.37 & 0.00$^\dagger$ & 139 \\
yes & 1.76 & 0.00$^\dagger$ & 4147 & options hmm & 2.08 & 0.00$^\dagger$ & 1107 & answer wait let & 1.54 & 0.14$^\dagger$ & 505 & trying figure answer question & 1.12 & 0.00$^\dagger$ & 203 \\
option & 1.75 & 0.00$^\dagger$ & 249743 & looking options & 1.62 & 0.00$^\dagger$ & 16721 & option perhaps answer & 1.48 & 0.14$^\dagger$ & 371 & figure option correctly describes & 1.12 & 0.00$^\dagger$ & 467 \\
answer & 1.46 & 0.00$^\dagger$ & 83503 & look options & 1.48 & 0.00$^\dagger$ & 1035 & options answer wait & 1.27 & 0.14$^\dagger$ & 159 & need figure answer question & 1.12 & 0.00$^\dagger$ & 208 \\
bit & 1.42 & 0.00$^\dagger$ & 11949 & options let & 1.34 & 0.00$^\dagger$ & 1894 & perhaps problem asking & 1.26 & 0.14$^\dagger$ & 361 & need figure option correct & 1.11 & 0.00$^\dagger$ & 151 \\
energy & 1.11 & 0.00$^\dagger$ & 2231 & option wait & 1.28 & 0.00$^\dagger$ & 6858 & answer alternatively perhaps & 1.25 & 0.14$^\dagger$ & 424 & figure correct answer question & 1.10 & 0.00$^\dagger$ & 996 \\
true & 1.11 & 0.00$^\dagger$ & 8331 & option let & 1.24 & 0.00$^\dagger$ & 868 & check options options & 1.19 & 0.14$^\dagger$ & 214 & options answer wait let & 1.09 & 0.00$^\dagger$ & 83 \\
say & 0.96 & 0.00$^\dagger$ & 54978 & question woman & 1.21 & 0.00$^\dagger$ & 470 & wait wait let & 1.18 & 0.14$^\dagger$ & 178 & need choose correct option & 1.09 & 0.00$^\dagger$ & 125 \\
sound & 0.95 & 0.00$^\dagger$ & 5481 & wait options & 1.20 & 0.00$^\dagger$ & 3318 & need understand option & 1.10 & 0.14$^\dagger$ & 255 & need figure right answer & 1.09 & 0.00$^\dagger$ & 92 \\
salary & 0.89 & 0.00$^\dagger$ & 486 & question happens & 1.20 & 0.00$^\dagger$ & 575 & answer wait wait & 1.08 & 0.14$^\dagger$ & 194 & looking options option talks & 1.09 & 0.00$^\dagger$ & 1614 \\
\cmidrule(lr){1-16}
\rowcolor{red!10}
\multicolumn{16}{c}{\textbf{Bottom 10 $n$-grams that reduce the confidence.}} \\
\cmidrule(lr){1-16}
denote & -5.99 & 0.00$^\dagger$ & 1379 & really stuck & -8.76 & 0.00$^\dagger$ & 865 & \textcolor{blue}{\textbf{equation let number}} & \textcolor{blue}{\textbf{-7.35}} & \textcolor{blue}{\textbf{0.14}}$^\dagger$ & \textcolor{blue}{\textbf{109}} & set equation let number & -6.04 & 0.00$^\dagger$ & 105 \\
consider & -5.82 & 0.00$^\dagger$ & 6602 & squared divisible & -4.50 & 0.00$^\dagger$ & 15 & \textcolor{blue}{\textbf{really stuck think}} & \textcolor{blue}{\textbf{-5.74}} & \textcolor{blue}{\textbf{0.14}}$^\dagger$ & \textcolor{blue}{\textbf{383}} & matching options wait perhaps & -5.14 & 0.00$^\dagger$ & 136 \\
perhaps & -3.89 & 0.00$^\dagger$ & 91550 & pythagorean theorem & -3.56 & 0.00$^\dagger$ & 36 & \textcolor{blue}{\textbf{scored points total}} & \textcolor{blue}{\textbf{-5.64}} & \textcolor{blue}{\textbf{0.14}}$^\dagger$ & \textcolor{blue}{\textbf{2}} & think ve exhausted possibilities & -4.03 & 0.00$^\dagger$ & 179 \\
subtract & -3.39 & 0.00$^\dagger$ & 2359 & divide total & -3.56 & 0.00$^\dagger$ & 123 & \textcolor{blue}{\textbf{km upstream km}} & \textcolor{blue}{\textbf{-5.52}} & \textcolor{blue}{\textbf{0.14}}$^\dagger$ & \textcolor{blue}{\textbf{44}} & time distance divided speed & -4.00 & 0.00$^\dagger$ & 20 \\
stuck & -3.13 & 0.00$^\dagger$ & 4782 & km upstream & -3.45 & 0.00$^\dagger$ & 83 & \textcolor{blue}{\textbf{student taken subjects}} & \textcolor{blue}{\textbf{-4.95}} & \textcolor{blue}{\textbf{0.14}}$^\dagger$ & \textcolor{blue}{\textbf{2}} & downstream km upstream km & -3.88 & 0.00$^\dagger$ & 35 \\
listed & -2.87 & 0.00$^\dagger$ & 4093 & degree degree & -3.27 & 0.00$^\dagger$ & 37 & \textcolor{blue}{\textbf{complete work days}} & \textcolor{blue}{\textbf{-3.85}} & \textcolor{blue}{\textbf{0.14}}$^\dagger$ & \textcolor{blue}{\textbf{54}} & approach differently let consider & -3.75 & 0.00$^\dagger$ & 10 \\
acre & -2.40 & 0.00$^\dagger$ & 39 & power source & -3.05 & 0.00$^\dagger$ & 30 & \textcolor{blue}{\textbf{mentions million people}} & \textcolor{blue}{\textbf{-3.85}} & \textcolor{blue}{\textbf{0.14}}$^\dagger$ & \textcolor{blue}{\textbf{4}} & largest positive integer divide & -3.27 & 0.00$^\dagger$ & 3 \\
problem & -1.98 & 0.00$^\dagger$ & 25482 & total parts & -2.91 & 0.00$^\dagger$ & 181 & \textcolor{blue}{\textbf{total parts parts}} & \textcolor{blue}{\textbf{-3.75}} & \textcolor{blue}{\textbf{0.14}}$^\dagger$ & \textcolor{blue}{\textbf{33}} & really stuck think correct & -2.78 & 0.00$^\dagger$ & 93 \\
sp & -1.86 & 0.00$^\dagger$ & 2513 & think ve & -2.74 & 0.00$^\dagger$ & 965 & \textcolor{blue}{\textbf{cork didn fit}} & \textcolor{blue}{\textbf{-3.02}} & \textcolor{blue}{\textbf{0.14}}$^\dagger$ & \textcolor{blue}{\textbf{4}} & selling price selling price & -2.68 & 0.00$^\dagger$ & 20 \\
total & -1.83 & 0.00$^\dagger$ & 19905 & let problem & -2.56 & 0.00$^\dagger$ & 537 & \textcolor{blue}{\textbf{different points time}} & \textcolor{blue}{\textbf{-2.96}} & \textcolor{blue}{\textbf{0.14}}$^\dagger$ & \textcolor{blue}{\textbf{3}} & correct answer options perhaps & -2.60 & 0.00$^\dagger$ & 194 \\
    \bottomrule
    \end{tabular}
    }
    \caption{
    Confidence analysis using Lasso and accuracy analysis using logistic for R1 Qwen.
    We apply Z-score standardization to the $n$-gram scores across all features.
    $^\dagger$ indicates results where the original feature value was 0 before standardization.
    \textcolor{blue}{\textbf{Blue bold text}} indicates features for which confidence decreased, and accuracy increased.
    }
    \label{tab:lasso-result-all-1}
\end{table*}

\paragraph{$N$-grams Influencing Overconfidence.}
Table~\ref{tab:lasso-result-all-1} lists the top 10 $n$-grams with the largest positive regression weights~(i.e., linguistic expressions increasing confidence) and the bottom 10 with the largest negative weights~(i.e., decreasing confidence) normalized by the Z-score standardization to allow comparison across different $n$-gram sizes.

In Table~\ref{tab:lasso-result-all-1}, $n$-grams identified to boost confidence include many expressions that suggest multiple possibilities without finalizing an answer, such as ``option,'' ``option hmm let try,'' ``maybe,'' ``perhaps,'' and ``perhaps wait perhaps calculate.''
In contrast, expressions identified to cause a drop in confidence include many terms for organizing thoughts or re-evaluating, such as ``perhaps mistake wait let check,'' ``let check,'' ``subtract,'' ``make sure didn't,'' and ``make mistake.''
In addition, previous research~\citep{muennighoff-etal-2025-s1} shows that intentionally inserting words like ``wait'' or ``hmm'' to lengthen reasoning contributes to performance gains, known as ``test-time scaling'' or ``slow thinking''~\citep{gandhi2025cognitive,min2024imitate}.
Our experimental result clarifies that these words are not only vital for improving performance but also act as key factors affecting model confidence.

Furthermore, Table~\ref{tab:correlation-bw-generation-and-forceddecoding} presents the correlation of extracted $n$-grams between the generation-based and forced-decoding methods when measuring the confidence, evaluated by Pearson correlation and Jaccard index ($k$=50, 100).
For \{1--3\}-grams but for 5-gram, the correlation reaches as high as 0.85, indicating that the $n$-grams identified through regression analysis are largely consistent across different methods, strengthening our findings.

\paragraph{$N$-grams Influencing Underconfidence.}
We find that the $n$-grams associated with model underconfidence do not necessarily align with those humans would intuitively expect.
From a psychological perspective~\cite{clark2002using}, humans in uncertain or confusing situations are known to express phrases such as ``uh, '' ``um'', ``I don't know'' or ``I am unsure.''
However, compared with Table~\ref{tab:lasso-result-all-1}, such expressions do not appear among the identified $n$-grams, indicating two findings: (1) the $n$-grams produced when models are underconfident differ from those observed in human psychology, and (2) these confidence-related $n$-grams act as \textit{superficial cues}, since their influence on model confidence is not readily apparent from a human perspective.
This highlights the importance of identifying them through regression-based analysis.

\begin{figure*}[t]
    \centering
    \includegraphics[width=\textwidth]{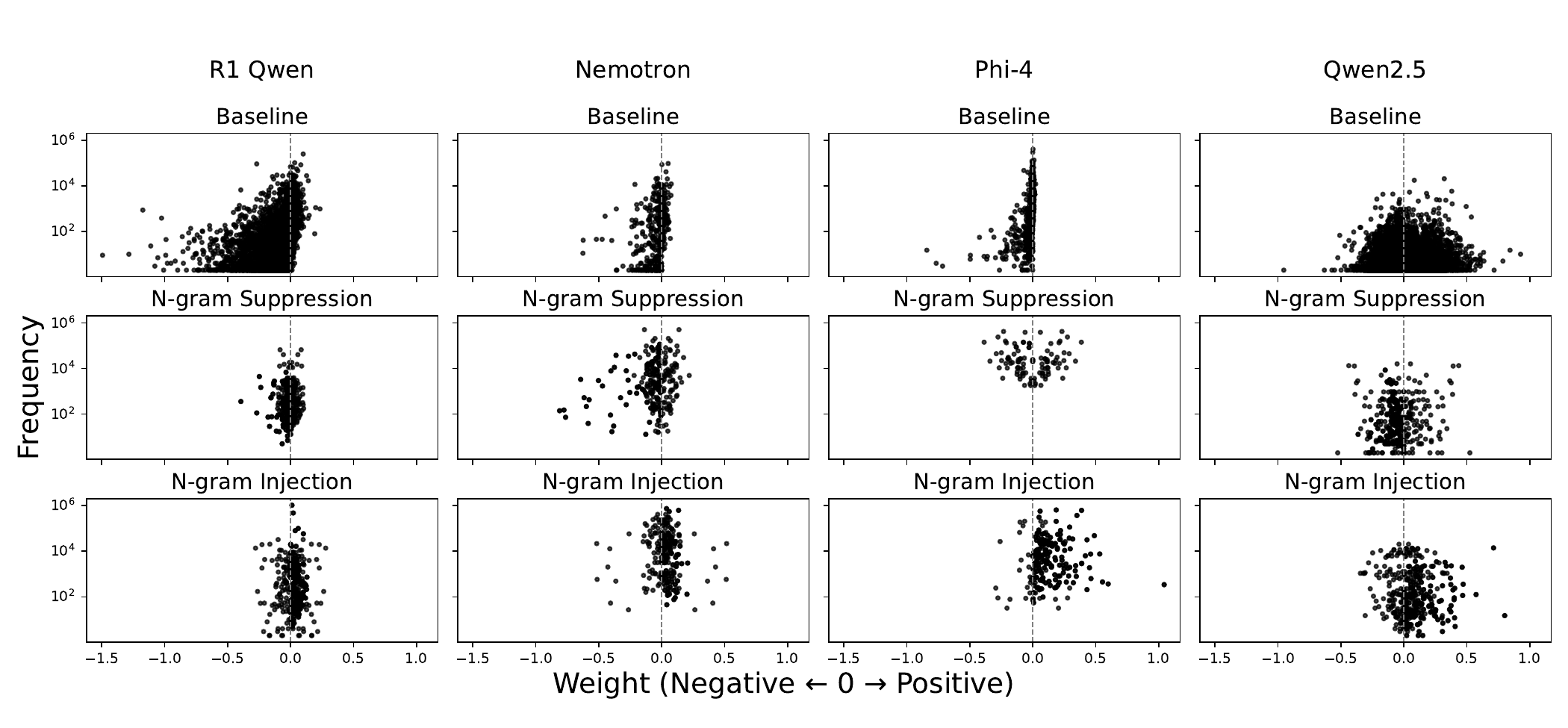}
    \caption{Changes in $n$-gram frequencies.}
    \label{fig:freq_change_rate}
\end{figure*}

\paragraph{Causal Effect Between $N$-grams and Confidence.}
\label{causal-relationship}
While the identified $n$-grams may correlate with overconfidence, $n$-grams might not be the sole cause, i.e., other factors could cause overconfidence~\citep{ma2026sparse}, as we discussed in \S\ref{re-generation}.
However, the successful manipulation results for $N$-gram Suppression in Table~\ref{tab:overall-results} alleviate this concern by providing evidence for a causal relationship between the identified $n$-grams and model confidence.
We also present the $N$-gram Injection results in Table~\ref{tab:overall-results}, in which confidence is intentionally deteriorated, leading to the expected result, i.e., the models became more overconfident.

As for the result of $N$-gram Suppression, we observe mitigated overconfidence in many models while maintaining accuracy levels; for instance, ACE decreases from 0.198 to 0.166 for R1 Qwen3 and from 0.415 to 0.334 for Nemotron by $N$-gram Suppression.
Since Qwen2.5 and Phi-4 mini also show the same tendency as reasoning models, we can conclude that many models possess $n$-grams that impact the degree of confidence.
Moreover, besides calibration metrics, the drop in Brier score for many models indicates a reduction in the error of the probability predictions themselves.
The AUC-ROC (AUROC) also improves in many models, showing that our method enhances the discriminative capability in confidence (i.e., the ability to distinguish between correct and incorrect answers).

\begin{figure*}[t]
    \centering
    \includegraphics[width=\linewidth]{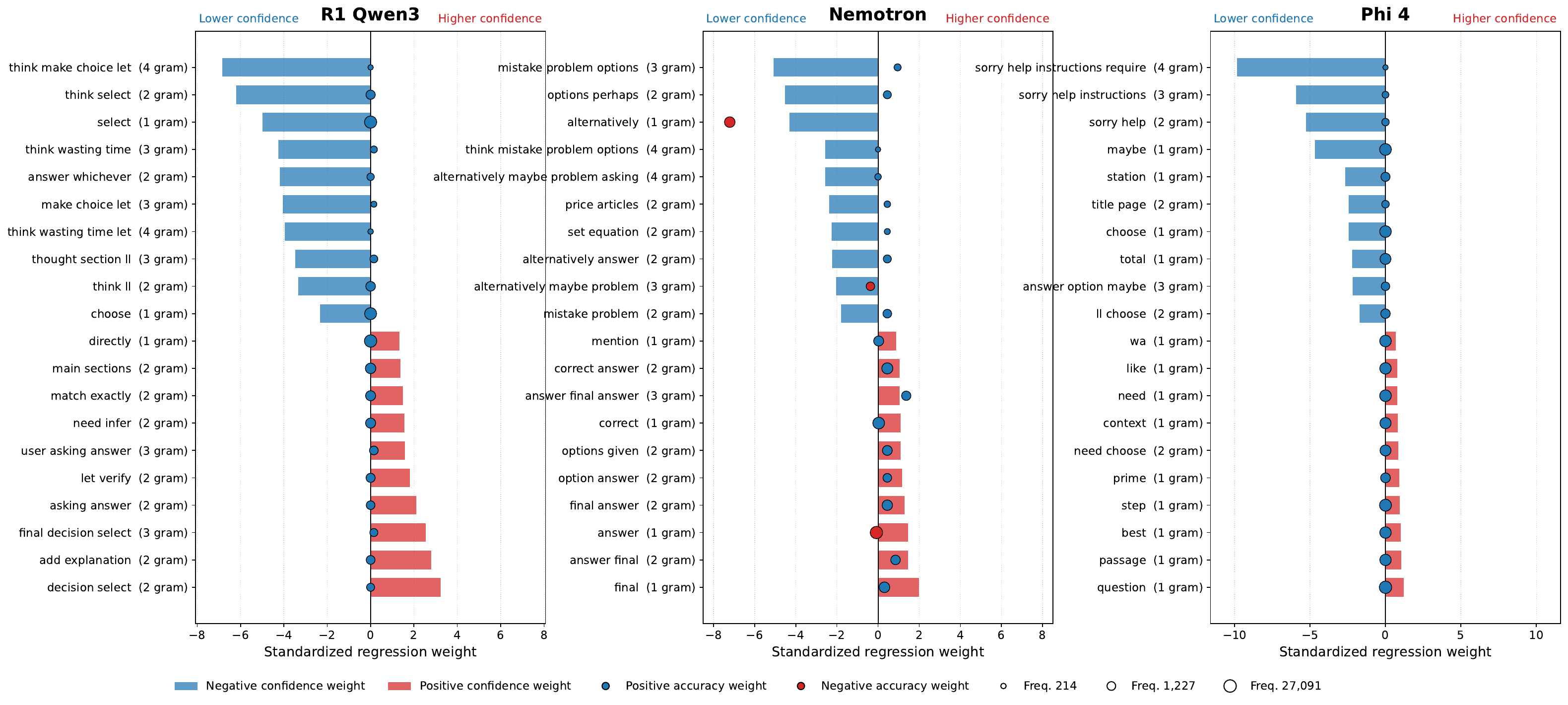}
    \caption{
    Confidence-related $n$-grams for three reasoning models.
    Bars show standardized regression weights on confidence, and circles show standardized regression weights on accuracy.
    Circle size indicates $n$-gram frequency.
    }
    \label{fig:divergining_bar}
\end{figure*}

\begin{figure*}[t]
    \centering
    \includegraphics[width=\textwidth]{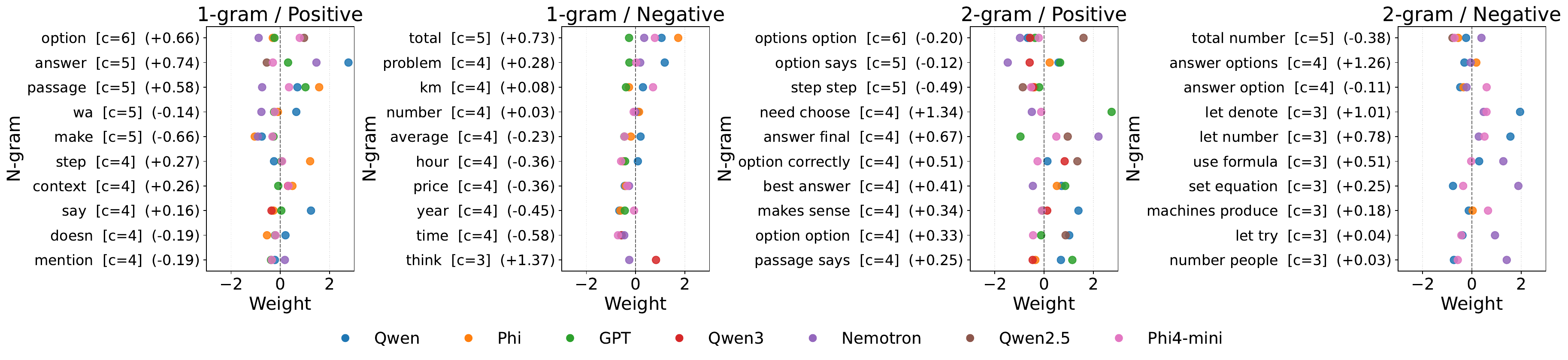}
    \caption{Results of extracting $n$-grams common to the models.}
    \label{fig:generalized-features}
\end{figure*}

Figure~\ref{fig:freq_change_rate} visualizes how the $n$-grams appearing in the Baseline change during $N$-gram Suppression/Injection.
Each data point represents the corresponding weight and frequency in each Re-generation for a fixed set of $n$-grams that appeared with high frequency in the Baseline.
In $N$-gram Suppression, we observe a relatively high number of $n$-grams with negative weights.
Conversely, $N$-gram Injection shows an increase in $n$-grams with positive weights, justifying the frequency shift.

\paragraph{Influential $N$-grams on Both Acc. and Conf.}
Table~\ref{tab:lasso-result-all-1} compares the $n$-grams identified by Lasso regression for confidence and logistic regression for accuracy.
The entries highlighted in \textbf{\textcolor{blue}{blue}} correspond to $n$-grams that decrease confidence while improving prediction accuracy.
Although such expressions exist, they appear only sporadically across models.
For example, in R1 Qwen, expressions such as ``correct answer wait let'' are associated with both lower confidence and higher accuracy, suggesting that they encourage more careful reasoning.
However, these examples represent only a small subset of the extracted expressions, indicating that confidence-related and accuracy-related linguistic cues only partially overlap.

Figure~\ref{fig:divergining_bar} further illustrates this relationship by comparing the standardized regression weights for confidence (bars) and accuracy (circles).
Across all three reasoning models, the bars and circles are frequently misaligned, showing that linguistic expressions strongly associated with confidence are often weakly related to prediction accuracy.
For example, several influential $n$-grams in R1 Qwen3 and Nemotron have large confidence weights while their corresponding accuracy weights remain close to zero.
Conversely, only a limited number of expressions exhibit consistently positive effects on both confidence and accuracy.
These observations suggest that confidence-related and accuracy-related expressions arise from distinct linguistic factors, supporting our hypothesis that superficial linguistic cues can substantially influence model confidence without reflecting actual correctness.%
\footnote{
We also perform regression analyses on 6-, 7-, and 8-grams and observe similar trends.
To compensate for the lower frequency of longer $n$-grams, we additionally replace low-frequency words with POS tags~\cite{bisazza-federico-2012-cutting}, obtaining consistent observations.
}

\paragraph{Cross-Model Confidence-Related $N$-grams.}
Figure~\ref{fig:generalized-features} shows $n$-grams shared across multiple LLMs and their impacts on confidence in each model.
The x-axis represents $z$-score standardized weights, where points further to the right indicate stronger positive correlations with confidence.
The y-axis lists the $n$-grams, where $c$ in parentheses denotes the number of models in which each $n$-gram appears (e.g., $c = 4$ indicates that the $n$-gram appears across 4 models).

Our analysis reveals that shared confidence-related patterns emerge in shorter $n$-grams (i.e., 1- and 2-grams), while longer $n$-grams are rarely shared across models.
For 1- and 2-grams, several confidence-related expressions consistently appear across multiple models.
On the positive side, expressions such as ``final answer'' are frequently shared, whereas uncertainty-related expressions such as ``mistake'' and ``perhaps'' appear on the negative side, suggesting that LLMs share common lexical cues associated with confidence.

Interestingly, Figure~\ref{fig:generalized-features} also reveals that several shared expressions correspond to degenerate generation patterns rather than meaningful reasoning.
For instance, repetitive $n$-grams such as ``options option,'' ``option option,'' and ``step step'' appear across multiple models and are associated with confidence.
Since these repetitions carry little semantic information for solving the task, their influence on confidence suggests that LLMs partly rely on superficial generation cues rather than reasoning quality when forming confidence estimates.%
\footnote{
Appendix~\ref{app:reasoning-length-bias} also examines whether reasoning length affects confidence and finds that it does not affect confidence.
}

\begin{table}[t]
    \centering
    \resizebox{\linewidth}{!}{
    \setlength{\tabcolsep}{2pt}    
    \begin{tabular}{c ccccc}
    \toprule
    \textbf{Metric} & \textbf{1-gram} & \textbf{2-gram} & \textbf{3-gram} & \textbf{4-gram} & \textbf{5-gram} \\
    \midrule
    \rowcolor{yellow!10}
    \multicolumn{6}{l}{\textbf{P(True) (\citet{kadavath2022languagemodelsmostlyknow})}} \\
    \midrule    
    Pearson & 0.73 & 0.68 & 0.63 & 0.69 & 0.54 \\
    Jaccard@50 & 0.88 & 0.86 & 0.87 & 0.84 & 0.81 \\
    Jaccard@100 & 0.93 & 0.91 & 0.92 & 0.90 & 0.88 \\
    \midrule
    \rowcolor{green!10}
    \multicolumn{6}{l}{\textbf{Seq. Prob. (\citet{jiang-etal-2021-know,lin-etal-2024-contextualized})}} \\
    \midrule
    Pearson & 0.61 & 0.68 & 0.53 & 0.59 & 0.34 \\
    Jaccard@50 & 0.72 & 0.69 & 0.64 & 0.56 & 0.50 \\
    Jaccard@100 & 0.85 & 0.81 & 0.80 & 0.73 & 0.69 \\
    \bottomrule
    \end{tabular}
    }
    \caption{Correlation between $n$-grams from open-ended generation and MCQA tasks. @$k$ indicates the top-$k$.}
    \label{tab:correlation-bw-open-ended-generation-and-mcqa}
\end{table}

\paragraph{Correlation with Open-ended Generation Task.}
Our main experiments analyze the $n$-grams that influence model confidence using multiple-choice question answering tasks. 
To investigate whether similar confidence-related $n$-grams can also be identified in open-ended generation tasks, we evaluate 5 generation benchmarks: GSM8k~\cite{cobbe2021trainingverifierssolvemath}, WMT  (2019~\cite{barrault-etal-2019-findings} and 2023~\cite{kocmi-etal-2023-findings}), XSum~\cite{narayan-etal-2018-dont}, and SAMSum~\cite{gliwa-etal-2019-samsum}, comprising a total of 18,492 instances.
Following P(True) by \citet{kadavath2022languagemodelsmostlyknow}, we estimate confidence in open-ended generation by measuring the generation probability of whether the produced answer is true or false.
To further strengthen the validity, we also use the likelihood of the generated output~\cite{jiang-etal-2021-know,lin-etal-2024-contextualized}, defined as Seq. Prob., as an additional confidence score.
Appendix~\ref{app:analysis-on-open-ended-generation-qa-tasks} provides the detailed notations.

Table~\ref{tab:correlation-bw-open-ended-generation-and-mcqa} shows the correlation between the confidence-related $n$-grams extracted from open-ended generation tasks and those extracted from MCQA tasks.
The results show consistently high correlations across all $n$-gram lengths, suggesting that the linguistic patterns affecting model confidence are largely shared across task formats.

\begin{table}[t]
    \centering
    \small
    \begin{tabular}{l rrr}
        \toprule
        \textbf{Model} & \textbf{$\Delta$ ECE} & \textbf{$\Delta$ Acc.} & \textbf{$\Delta$ AUROC} \\
        \midrule
        R1 Qwen    & -0.018 & +0.006 & -0.011 \\
        GPT-oss    & -0.004 & +0.014 & +0.017 \\
        Phi-4      & -0.011 & +0.042 & +0.096 \\
        Qwen2.5    & -0.103 & +0.184 & -0.157 \\
        Phi-4 mini & -0.131 & +0.008 & -0.176 \\
        \bottomrule
    \end{tabular}
    \caption{Leave-two-out results using $N$-grams extracted from RACE, CosmosQA, and HellaSwag, evaluated on the held-out MathQA and MMLU datasets.}
    \label{tab:leave-two-out-results}
\end{table}

\begin{figure*}[t]
    \centering
    \includegraphics[width=\textwidth]{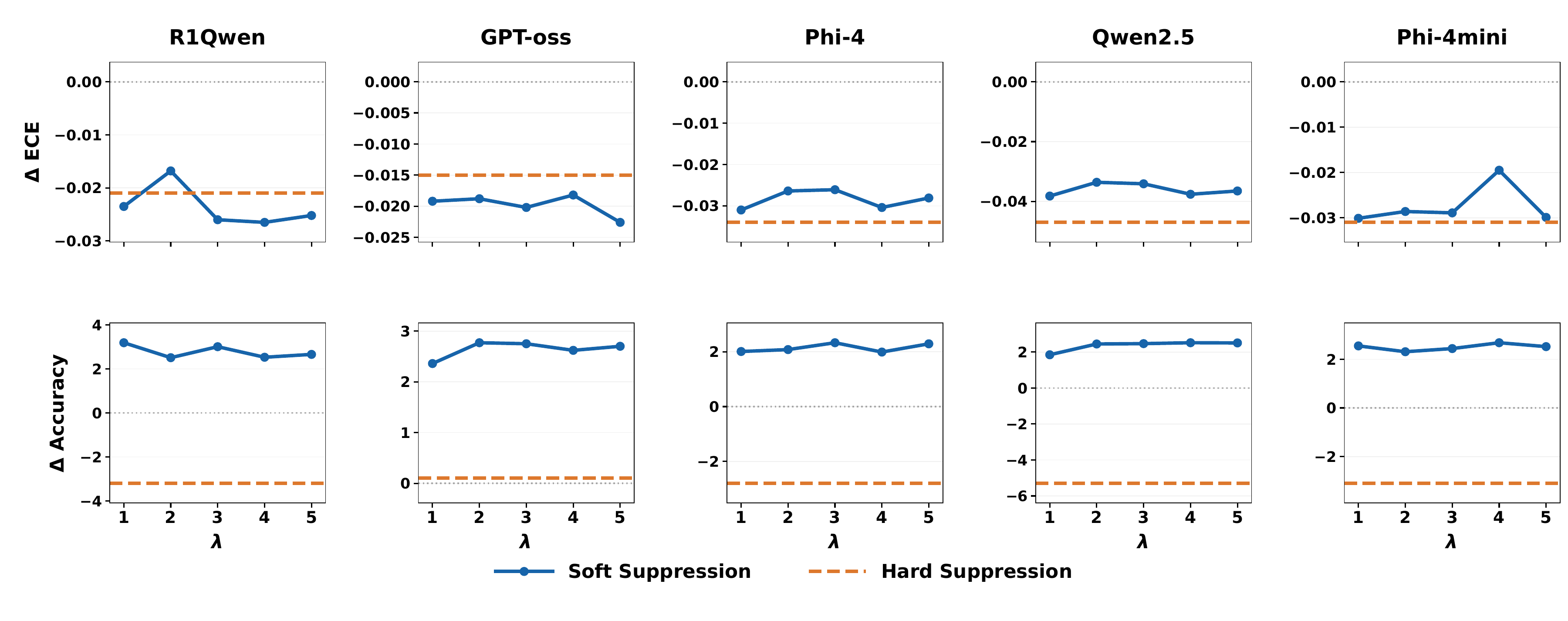}
    \caption{Effects of N-gram Suppression using constrained decoding. We compare Soft Suppression and Hard Suppression across five models. The x-axis represents the penalty strength $\lambda$ for Soft Suppression, while the y-axis shows changes in ECE (top) and accuracy (bottom) relative to the baseline.}
    \label{fig:const_decoding}
\end{figure*}

\paragraph{\red{Leave-two-out Ablation.}}
\red{
To investigate whether the identified confidence-related n-grams generalize across datasets, we conduct a leave-two-out ablation study on five models.
Specifically, we extract confidence-increasing n-grams from RACE, CosmosQA, and HellaSwag, and apply $N$-gram Suppression to the held-out MathQA and MMLU datasets.
Table~\ref{tab:leave-two-out-results} presents the results and shows improvements for three of the five models, with particularly notable gains in Qwen2.5 and Phi-4 mini.
Moreover, accuracy improves across all five models, suggesting that suppressing n-grams extracted from other datasets can be mitigated.
These findings show that confidence-related n-grams are not dataset-specific, but instead capture cues that generalize across different downstream tasks.
}

\paragraph{\red{$N$-gram Suppression by Constrained Decoding.}}
\red{
We employ prompting-based interventions to suppress confidence-increasing n-grams.
To examine the generalizability of $N$-gram Suppression beyond prompting, we additionally explore constrained decoding as an alternative intervention strategy~\cite{lu-etal-2021-neurologic}. 
Specifically, we adopt two approaches: (1) \underline{Hard Suppression}, which completely prohibits the generation of identified n-grams by setting their corresponding logits to $-\infty$, and (2) \underline{Soft Suppression}, which subtracts a penalty $\lambda$ from the corresponding logits during decoding, i.e., $\text{logits} \leftarrow \text{logits} - \lambda$. We vary the penalty strength $\lambda \in \{1, 2, 3, 4, 5\}$.
}

\red{
Figure~\ref{fig:const_decoding} presents the changes in ECE and accuracy under different penalty strengths. 
We observe that Soft Suppression reduces ECE in several models while generally maintaining accuracy. 
In contrast, Hard Suppression does not consistently improve calibration and can lead to substantial decreases in accuracy, particularly for Qwen2.5. 
This suggests that completely prohibiting the generation of identified n-grams may constrain the reasoning process, thereby degrading prediction performance. 
These results indicate that Soft Suppression provides a more flexible intervention by discouraging confidence-increasing n-grams without completely eliminating them. 
Furthermore, the stable accuracy under Soft Suppression suggests that model confidence can be adjusted without degrading prediction performance, providing additional evidence that the identified n-grams influence confidence.
}

\section{Conclusion}
Focusing on what $n$-grams affect confidence, we extract $n$-grams that influence confidence through regression on reasoning segments.
Our results show that specific expressions affect confidence, some of which align with those used for test-time scaling.
We also confirm their causal relationship on model confidence by suppressing the identified $n$-grams during reasoning, demonstrating that manipulating these linguistic cues changes model confidence, providing evidence that they affect confidence formation.
Lastly, we show that our method applies not only to MCQA task but also to open-ended generation.

\clearpage
\section*{Limitations}
Our study admits that several limitations remain.

(1) \textbf{Proprietary models.}
Gemini Thinking exists as a proprietary model and as an API model that allows access to reasoning processes via the \texttt{<think>} tag~\citep{comanici2025gemini}.
However, running the same experiments on this model would incur a cost of around \$0.39 USD per question $\times$ 21,500 instances $\times$ 3 settings $=$ \$25,155 USD.
This cost far exceeds our budget, so we did not conduct the experiments.
Our paper also does not study the OpenAI o1/o3 models~\citep{jaech2024openai} as well, because we do not have access to their chain-of-thought, and thus, cannot evaluate those.

(2) \textbf{Scope.}
Our work is not primarily positioned within the literature on confidence calibration.
Our main contribution is to investigate ``\textit{why models become overconfident}'' and ``\textit{what linguistic factors affect model confidence}'' as introduced in the Introduction.
Specifically, we identify a causal relationship between linguistic expressions and model confidence. 
As a result, our approach also leads to modest improvements in confidence calibration.
Our objective is not to outperform existing calibration methods, but rather to reveal the underlying causal mechanisms behind model overconfidence. Thus, we believe that direct comparison with existing calibration approaches is not essential to the central contribution of our work.

(3) \textbf{Context.}
The contribution of an $n$-gram to model confidence may depend on its surrounding context.
To mitigate this concern, our approach analyzes confidence by decomposing reasoning traces into $n$-grams, focusing on localized linguistic signals rather than full-context representations.
Therefore, we believe our research direction is fundamentally different from context-based analyses.
To partially account for contextual effects, we additionally conducted experiments using longer $n$-grams (up to 8-grams).
Since these experiments showed trends similar to those observed with shorter $n$-grams, our results suggest that confidence-related linguistic patterns may be relatively robust to local context.
However, this observation is not sufficient to conclude that such patterns are entirely context-independent.
In addition, to increase the data size for regression analysis, we aggregate five QA benchmarks in our main experiments.
While confidence-related linguistic patterns may differ across tasks or domains, many task-specific $n$-grams were not assigned substantial weights due to the sparsity induced by Lasso regression.
Collecting larger domain-specific QA datasets may enable more fine-grained analysis of task-dependent confidence signals.
We leave for future work the investigation, i.e., (1) task-specific $n$-grams across different domains and (2) how the contribution/meaning of an $n$-gram changes depending on the context.

(4) \textbf{Other Intervention Methods.}
In this work, we adopt prompting-based intervention methods as introduced in \S\ref{re-generation}, namely ``$N$-gram Suppression'' and ``$N$-gram Injection'', to investigate the causal effect of the identified $n$-grams on model confidence.
However, alternative intervention strategies, such as constrained decoding, could also be considered~\cite{ballon2026probingtrajectoriesreasoningtraces,pmlr-v235-beurer-kellner24a}.
Our primary goal is to verify whether the identified $n$-grams truly influence confidence, and thus we employ prompting-based methods due to their simplicity and reproducibility.
Investigating whether similar causal effects hold under other decoding-based intervention approaches remains an important direction for future work, as we do not introduce a brand-new calibration approach.

(5) \textbf{Influential $N$-grams across Models.}
Although we identified several shared confidence-related patterns across models, these common patterns were primarily limited to shorter $n$-grams, while longer shared expressions were rarely observed, as discussed in \S\ref{results-and-discussions}.
Moreover, even when the same $n$-grams appeared across multiple models, their contributions to confidence varied substantially in magnitude and sometimes in direction.
Our approach is designed to identify confidence-related $n$-grams for each model individually, with the goal of uncovering model-specific linguistic patterns associated with confidence estimation.
Given differences in model architectures, training data, and alignment procedures, such variation is expected.

\section*{Ethical Considerations}
All used datasets were verified to be permissible for research purposes. 
Since all datasets are publicly available, our experiments are reproducible. 
Our work utilizes open-weight models through Hugging Face Transformers library~\cite{wolf-etal-2020-transformers} and lists all of detailed model names in Table~\ref{tab:detailed-model-name} in Appendix~\ref{detailed-experimental-setings}, confirming that those models are able to be used for research purposes.
Our work also uses AI tools, including LLMs (especially ChatGPT) and coding agents (Copilot), to conduct experiments and improve the quality of English writing.

\clearpage
\bibliography{custom}

\appendix
\section*{Appendix}
\section{Detailed Dataset Names and Subsets}
\label{detailed-dataset}
We list the detailed dataset names used in Table~\ref{tab:detailed-dataset-name}, including MCQAs and open-ended generation QAs.
In our study, we restrict the experiments to subsets of MMLU that require reasoning, i.e., STEM fields, and list the detailed subsets used in Table~\ref{tab:mmlu-subset}.
We use the validation set in HellaSwag~\cite{zellers-etal-2019-hellaswag}, since the test set has not been released yet.

\begin{table}[t]
    \centering
    \setlength{\tabcolsep}{2pt}
    \resizebox{\linewidth}{!}{
    \begin{tabular}{lccc}
    \toprule
    \textbf{QA} & \textbf{Data} & \textbf{HuggingFace} & \textbf{Citation} \\
    \midrule
    \rowcolor{yellow!10}
    \multicolumn{4}{l}{\textbf{Multiple-Choices QA}} \\
    \midrule
    HellaSwag & 10,042 & Rowan/hellaswag & \citet{zellers-etal-2019-hellaswag} \\
    MathQA & 2,985 & allenai/math\_qa & \citet{amini-etal-2019-mathqa} \\
    MMLU (STEM) & 1,990 & cais/mmlu & \citet{hendryckstest2021} \\
    RACE & 2,985 & EleutherAI/race & \citet{lai-etal-2017-race} \\
    CosmosQA & 3,498 & allenai/cosmos\_qa & \citet{huang-etal-2019-cosmos} \\    
    \cdashline{1-4}
    \textbf{Total} & 21,500 & & \\
    \midrule
    \rowcolor{green!10}
    \multicolumn{4}{l}{\textbf{Open-ended Generation}} \\
    \midrule
    GSM8K & 1,319 & openai/gsm8k & \citet{cobbe2021trainingverifierssolvemath} \\
    WMT'19 (en-de) & 1,997 & wmt/wmt19 & \citet{barrault-etal-2019-findings} \\
    WMT'23 (en-de) & 3,023 & wmt/wmt23 & \citet{kocmi-etal-2023-findings} \\
    XSum & 11,334 & EdinburghNLP/xsum & \citet{narayan-etal-2018-dont} \\
    SAMSum & 819 & knkarthick/samsum & \citet{gliwa-etal-2019-samsum} \\
    \cdashline{1-4}
    \textbf{Total} & 18,492 & & \\
    \bottomrule
    \end{tabular}
    }
    \caption{Datasets used and their sizes.}
    \label{tab:detailed-dataset-name}
\end{table}

\begin{table}[t]
    \centering
    \small
    \begin{framed}
        \noindent abstract\_algebra, college\_computer\_science, college\_mathematics, 
        college\_physics, conceptual\_physics, elementary\_mathematics, formal\_logic, 
        high\_school\_computer\_science, high\_school\_mathematics, high\_school\_physics, high\_school\_statistics, 
        machine\_learning
    \end{framed}
    \caption{MMLU subset used in our paper.}
    \label{tab:mmlu-subset}
\end{table}

\begin{table}[t]
    \centering
    \small
    \renewcommand{\arraystretch}{1.2}
    \resizebox{\linewidth}{!}{
    \begin{tabular}{lrcc}
    \toprule
    \textbf{Model} & \textbf{Parameter} & \textbf{Detailed Name on HuggingFace} & \textbf{Citation} \\
    \midrule
    \rowcolor{blue!10}
    \multicolumn{4}{l}{\textbf{Reasoning Models}} \\
    R1 Qwen & 7B & deepseek-ai/DeepSeek-R1-Distill-Qwen-7B & \citet{guo2025deepseek} \\
    Nemotron & 8B & nvidia/Llama-3.1-Nemotron-Nano-8B-v1  & \citet{bercovich2025llama}\\
    R1 Qwen3 & 8B & deepseek-ai/DeepSeek-R1-0528-Qwen3-8B  & \citet{guo2025deepseek} \\
    Phi-4 & 15B & microsoft/Phi-4-reasoning & \citet{abdin2025phi} \\
    GPT-oss & 20B & openai/gpt-oss-20b & \citet{agarwal2025gpt} \\
    \midrule
    \rowcolor{red!10}
    \multicolumn{4}{l}{\textbf{Instruction-tuned Models}} \\
    Qwen 2.5 & 72B & Qwen/Qwen2.5-72B-Instruct & \citet{qwen2024qwen2}\\
    Phi-4 mini & 4B & microsoft/Phi-4-mini-instruct & \citet{abouelenin2025phi} \\
    \bottomrule
    \end{tabular}
    }
    \caption{Detailed model names used in the experiments.}
    \label{tab:detailed-model-name}
\end{table}

\begin{table*}[t]
    \centering
    \rowcolors{6}{gray!10}{white}
    \resizebox{\linewidth}{!}{
    \begin{tabular}{c cccccc cccccc cccccc}
    \toprule
    \multirow{2.25}{*}{\textbf{Model}} & 
    \multicolumn{6}{c}{\textbf{Baseline Inference}} & 
    \multicolumn{6}{c}{\textbf{$N$-gram Suppression}} & 
    \multicolumn{6}{c}{\textbf{$N$-gram Injection}} \\
    \cmidrule(lr){2-7}\cmidrule(lr){8-13}\cmidrule(lr){14-19}
    &
    \textbf{M} & \textbf{H} & \textbf{Ma} & \textbf{R} & \textbf{C} & \textbf{Avg.} &
    \textbf{M} & \textbf{H} & \textbf{Ma} & \textbf{R} & \textbf{C} & \textbf{Avg.} &
    \textbf{M} & \textbf{H} & \textbf{Ma} & \textbf{R} & \textbf{C} & \textbf{Avg.} \\
    \midrule
    \rowcolor{blue!10}
    \multicolumn{19}{l}{\textbf{Reasoning Models}} \\
    R1 Qwen      & 86.5 & 52.4 & 77.2 & 71.6 & 57.6 & 62.67 & 87.3 & 54.0 & 77.3 & 69.6 & 56.1 & 63.04 & 85.7 & 52.3 & 75.5 & 69.9 & 54.2 & 61.64 \\
    R1 Llama     & 79.5 & 62.2 & 68.2 & 77.9 & 25.2 & 61.97 & 29.1 & 26.5 & 21.0 & 22.1 & 25.2 & 25.10 & 26.1 & 60.9 & 44.5 & 47.4 & 28.7 & 48.77 \\
    Nemotron     & 74.0 & 40.6 & 72.4 & 67.9 & 46.6 & 51.04 & 62.8 & 37.8 & 51.0 & 55.0 & 58.9 & 55.25 & 63.3 & 41.5 & 54.3 & 62.7 & 51.1 & 49.56 \\
    R1 Qwen3     & 95.3 & 70.5 & 84.8 & 89.2 & 70.1 & 75.89 & 94.4 & 62.6 & 100.0 & 86.8 & 71.5 & 82.61 & 94.2 & 65.6 & 93.5 & 86.6 & 70.9 & 81.50 \\
    Phi-4        & 94.5 & 73.7 & 79.2 & 86.8 & 79.1 & 79.54 & 94.2 & 74.2 & 78.5 & 87.2 & 32.3 & 73.38 & 94.0 & 71.3 & 77.7 & 85.6 & 32.3 & 71.37 \\
    GPT-oss      & 92.2 & 74.3 & 78.4 & 81.9 & 78.0 & 78.27 & 94.9 & 74.6 & 77.9 & 85.8 & 80.8 & 79.55 & 94.4 & 73.6 & 77.8 & 85.7 & 81.4 & 79.08 \\
    \rowcolor{red!10}
    \multicolumn{19}{l}{\textbf{Instruction-tuned Models}} \\
    Llama 3.3    & 24.4 & 50.4 & 21.5 & 34.6 & 27.6 & 38.25 & 24.0 & 33.7 & 21.2 & 23.4 & 25.3 & 28.21 & 23.9 & 25.8 & 16.8 & 25.2 & 23.9 & 24.03 \\
    Qwen 2.5     & 24.4 & 31.9 & 20.8 & 24.0 & 24.7 & 27.40 & 24.5 & 72.7 & 46.6 & 24.6 & 24.8 & 50.13 & 28.1 & 35.2 & 24.1 & 23.8 & 24.9 & 29.72 \\
    Phi-4-mini   & 70.5 & 63.4 & 61.4 & 75.0 & 65.5 & 65.98 & 53.6 & 78.0 & 50.8 & 57.8 & 54.7 & 65.43 & 24.2 & 25.1 & 21.2 & 25.9 & 24.3 & 24.52 \\

    \bottomrule
    \end{tabular}
    }
    \caption{
    Experimental results on accuracy (as our work uses MCQA, we can calculate) and their average for solving five QA tasks.
    M, H, Ma, R, and C indicate MMLU, HellaSwag, MathQA, RACE, and CosmosQA, respectively.
    }
    \label{tab:accuracy}
\end{table*}

\section{Detailed Experimental Settings}
\label{detailed-experimental-setings}
\paragraph{Inference.}
In our study, we conducted the experiments using the Transformers library~\citep{wolf-etal-2020-transformers}.
For generations, we set max\_new\_tokens to 32,768 (10 when generating confidence), top\_p to 0.95, temperature to 0.6, and the seed to 42, following the default setting.
We used NVIDIA RTX A6000, RTX 6000 Ada Generation, and A100-SXM4-40GB GPUs.
Table~\ref{tab:detailed-model-name} lists the detailed names, and Table~\ref{tab:accuracy} reports accuracy performances.

\paragraph{Regression.}
In our study, we ran the experiments with scikit-learn~\citep{scikit-learn}, and conducted a 3-fold cross-validation for robust evaluation.
We selected the regularization coefficient $\lambda$ via cross-validation with three folds~\citep{stone1974cross}, and we chose the value that minimized the mean squared error on the validation data.
We used default settings for all other parameters.
Tables~\ref{tab:lasso-regression-evaluation-1} and ~\ref{tab:lasso-regression-evaluation-2} show the results on error.
LogSumExp is applied to avoid over/underflow in Equation~\ref{eq:confidence}.
\footnote{\url{https://en.wikipedia.org/wiki/LogSumExp}}

\paragraph{Logistic Regression to Accuracy.}
To verify that our analysis of confidence is not merely capturing signals of answer accuracy, i.e., the possibility that confidence signals track answer correctness, we conduct an analysis using logistic regression~\citep{cox1958regression} to predict correctness from textual features.
We convert each reasoning part into a vector $\mathbf{g}_i \in \mathbb{R}^{d}$ featuring $n$-grams that appear twice or more, forming the matrix $\mathbf{G} \in \mathbb{R}^{D \times d}$, applying L2 normalization to mitigate sentence-length bias.
The classification model to predict the accuracy is:
\begin{align}
    P(a_i = 1 \mid \mathbf{g}_i) &=  \sigma\!\left(\mathbf{w}^\top\frac{\mathbf{g}_i}{\|\mathbf{g}_i\|_2}\right), 
\end{align}
, where $a_i \in \{0,1\}$ is the $i$-th label, $\mathbf{w} \in \mathbb{R}^d$ is the vector of logistic regression coefficients, and $\sigma(z)$ is the logistic function, i.e., $1\; / \;  (1 + e^{-z})$.
After training, our work computes $\hat{p}_i = P(a_i = 1 \mid \mathbf{g}_i)$ and determines $\hat{a}_i = 1$, when $\hat{p}_i$ is greater than $0.5$.

\begin{figure*}[t]
    \centering
    \includegraphics[width=\textwidth]{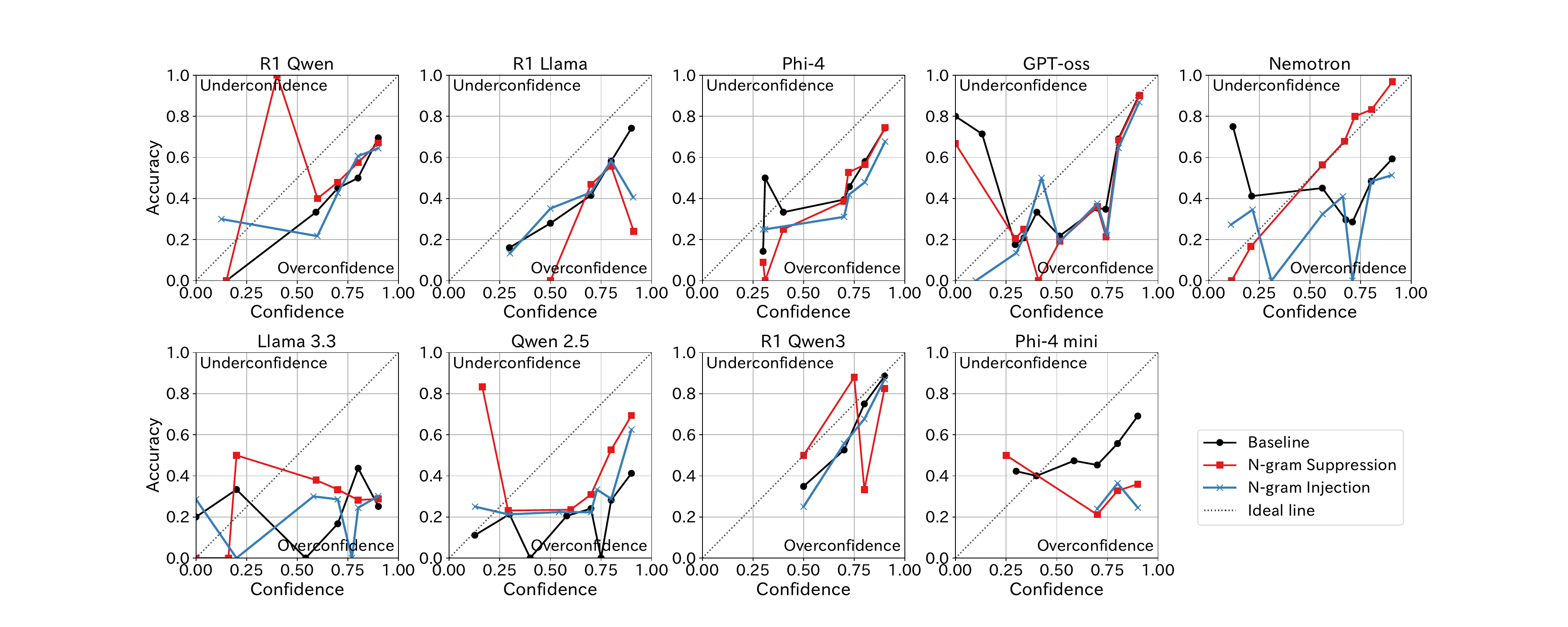}
    \caption{Calibration curve obtained using the generation-based method.}
    \label{fig:calibration_curve_generation}
\end{figure*}

\section{Detailed Evaluation Metrics}
\label{detailed-evaluation-metrics}
Our work adopts 4 evaluation metrics, i.e., Expected Calibration Error (ECE), Adaptive Calibration Error (ACE), Brier score, and AUC-ROC.
Each metric evaluates the validity of confidence estimation from a different viewpoint, which enables a comprehensive evaluation without solely relying on a single metric, as our study introduced in \S\ref{evaluation-metrics}.

Below, we explain each metric in detail with its formulation.
Here, $N$ denotes the number of evaluation samples, $\hat{p}_i \in [0,1]$ denotes the confidence output by the model for sample $i$, and $y_i \in {0,1}$ denotes the ground-truth label, where 1 indicates a correct answer, and 0 indicates an incorrect answer.

\paragraph{Expected Calibration Error (ECE).}
ECE~\citep{naeini2015obtaining} is a standard calibration error metric that measures the discrepancy between predicted confidence and actual accuracy.
Our work divides the confidence values $\hat{p}_i$ into $M$ bins (our work defines $M$ as $10$, following \citet{ozaki-etal-2025-understanding}), and defines ECE using the set of samples that belong to each bin $B_m$ as follows:
\begin{equation}
\mathrm{ECE} = \sum_{m=1}^{M} \frac{|B_m|}{N} \left| \mathrm{acc}(B_m) - \mathrm{conf}(B_m) \right|
\end{equation}
where $\mathrm{acc}(B_m) = \frac{1}{|B_m|} \sum_{i \in B_m} y_i$, and $\mathrm{conf}(B_m) = \frac{1}{|B_m|} \sum_{i \in B_m} \hat{p}_i$.
ECE is a lower-is-better metric, and values closer to 0 indicate more accurately calibrated confidence.

\paragraph{Adaptive Calibration Error (ACE).}
ACE~\citep{guo2017calibration} extends ECE and differs in that it constructs bins adaptively so that each bin contains the same number of samples.
This design mitigates statistical instability within each bin ($M$, and is equal to $10$) even when the confidence distribution is skewed.
The definition follows ECE.
We define:
\begin{equation}
\mathrm{ACE} = \frac{1}{M} \sum_{m=1}^{M} \left| \mathrm{acc}(B_m) - \mathrm{conf}(B_m) \right|
\end{equation}
As with ECE, ACE is a lower-is-better metric.

\paragraph{Brier Score.}
The Brier score~\citep{glenn1950verification} directly measures the probabilistic error of confidence predictions, and we define the Brier score:
\begin{equation}
\mathrm{Brier} = \frac{1}{N} \sum_{i=1}^{N} (\hat{p}_i - y_i)^2.
\end{equation}
Here, $\hat{p}_i$ indicates the prediction, $y_i$ is the gold in the instance $i$, and this metric reflects both calibration error and discriminative performance.
The Brier score is a lower-is-better metric, and it equals 0 for perfectly accurate confidence predictions.

\begin{table*}[t]
    \centering
    \small
    \begin{tabular}{lrrrrrrr}
        \toprule
        Metric & R1 Qwen3 & R1 Qwen & Nemotron & Phi-4 mini & Phi-4 & Qwen2.5 & GPT-oss \\
        \midrule
        $\Delta$ mean step score & $+0.0005$ & $+0.0050$ & $+0.0150$ & $-0.0079$ & $+0.0036$ & $-0.0085$ & $-0.0029$ \\
        \bottomrule
    \end{tabular}%
    \caption{Mean step scores evaluated by a process reward model.}
    \label{tab:prm-results}
\end{table*}

\paragraph{AUC-ROC.}
AUC-ROC measures the ability to discriminate between correct and incorrect answers using confidence scores.
Let $\hat{p}_i \in [0,1]$ denote the confidence score predicted by the model for the $i$-th example, and let $y_i \in \{0,1\}$ denote the ground-truth correctness label, where $y_i = 1$ indicates a correct prediction and $y_i = 0$ indicates an incorrect prediction.
Our work is treated as a binary classification task that uses $\hat{p}_i$ as the ranking.

AUC-ROC is defined as the area under the Receiver Operating Characteristic (ROC) curve as:
\begin{equation}
\mathrm{AUC} = \Pr(\hat{p}_{\text{pos}} > \hat{p}_{\text{neg}})
\end{equation}
where $\hat{p}_{\text{pos}}$ denotes the confidence score of a randomly selected positive example ($y=1$), and $\hat{p}_{\text{neg}}$ denotes the confidence score of a randomly selected negative example ($y=0$).
AUC-ROC is a higher-is-better metric.
A value of $0.5$ corresponds to random prediction, while $1.0$ indicates perfect.

\begin{figure*}[t]
    \centering
    \includegraphics[width=\textwidth]{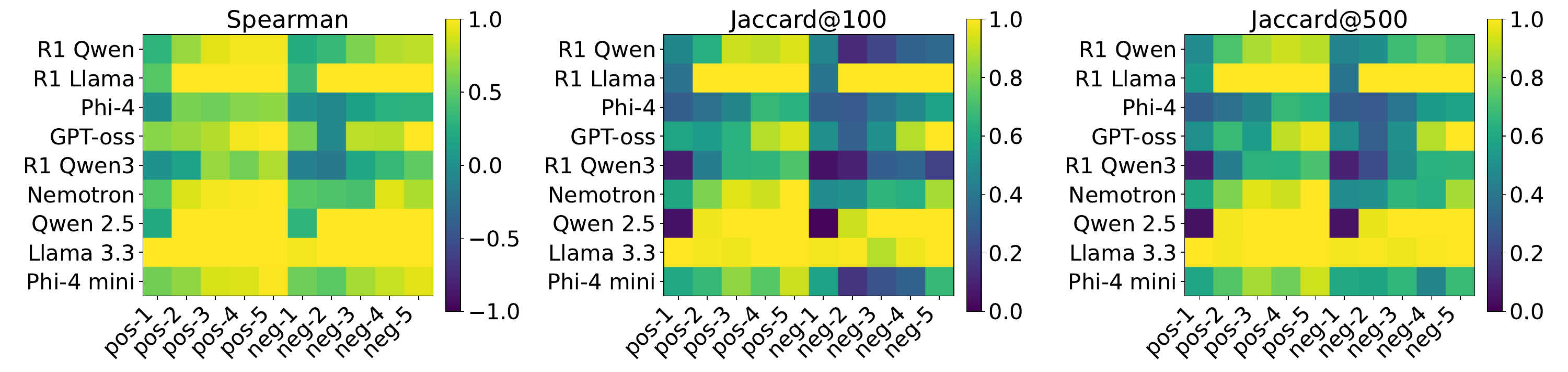}
    \caption{Correlation between TF-IDF-based and count-based methods.}
    \label{fig:correlation-between-different-features}
\end{figure*}

\begin{table*}[t]
    \centering
    \setlength{\tabcolsep}{1pt}
    \renewcommand{\arraystretch}{1.2}
    \rowcolors{6}{gray!10}{white}
    \resizebox{\linewidth}{!}{
    \begin{tabular}{c ccccc ccccc ccccc}
    \toprule
    \multirow{2.25}{*}{\textbf{Model}} & \multicolumn{5}{c}{\textbf{Baseline Inference}} & \multicolumn{5}{c}{\textbf{$N$-gram Suppression}} & \multicolumn{5}{c}{\textbf{Suppression only $N$-gram}} \\
    \cmidrule(lr){2-6}\cmidrule(lr){7-11}\cmidrule(lr){12-16}
     & \textbf{ECE $\downarrow$} & \textbf{ACE $\downarrow$} & \textbf{Brier $\downarrow$} & \textbf{Acc. $\uparrow$} & \textbf{AUROC $\uparrow$}
     & \textbf{ECE $\downarrow$} & \textbf{ACE $\downarrow$} & \textbf{Brier $\downarrow$} & \textbf{Acc. $\uparrow$} & \textbf{AUROC $\uparrow$}
     & \textbf{ECE $\downarrow$} & \textbf{ACE $\downarrow$} & \textbf{Brier $\downarrow$} & \textbf{Acc. $\uparrow$} & \textbf{AUROC $\uparrow$} \\
    \midrule
\rowcolor{blue!10}
\multicolumn{16}{l}{\textbf{Reasoning Models}} \\
\midrule
R1 Qwen & 0.334 & 0.334 & 0.713 & 0.627 & 0.502 & \textbf{0.318} & \textbf{0.316} & \textbf{0.700} & \textbf{0.630} & 0.510 & 0.323 & 0.323 & 0.708 & 0.625 & \textbf{0.516} \\
R1 Qwen3 & 0.198 & 0.198 & 0.450 & 0.759 & \textbf{0.604} & \textbf{0.166} & \textbf{0.166} & \textbf{0.344} & \textbf{0.826} & 0.566 & 0.169 & 0.170 & 0.361 & 0.820 & 0.566 \\
Nemotron & 0.415 & 0.415 & 0.896 & 0.510 & 0.517 & \textbf{0.334} & \textbf{0.334} & \textbf{0.859} & \textbf{0.553} & \textbf{0.637} & 0.404 & 0.404 & 0.907 & 0.476 & 0.551 \\
Phi-4 & 0.199 & 0.199 & \textbf{0.405} & \textbf{0.795} & 0.562 & \textbf{0.194} & \textbf{0.194} & 0.448 & 0.734 & \textbf{0.689} & 0.201 & 0.200 & 0.408 & 0.794 & 0.567 \\
GPT-oss & 0.191 & 0.191 & 0.412 & 0.783 & 0.532 & 0.191 & 0.191 & 0.401 & 0.795 & \textbf{0.546} & \textbf{0.188} & \textbf{0.188} & \textbf{0.398} & \textbf{0.797} & 0.525 \\
\midrule
\rowcolor{red!10}
\multicolumn{16}{l}{\textbf{Instruction-tuned Models}} \\
\midrule
Qwen 2.5 & 0.329 & 0.329 & 0.944 & 0.274 & 0.587 & \textbf{0.205} & \textbf{0.205} & \textbf{0.657} & \textbf{0.501} & \textbf{0.777} & 0.287 & 0.286 & 0.821 & 0.391 & 0.642 \\
Phi-4 mini & 0.285 & 0.283 & \textbf{0.640} & \textbf{0.660} & 0.509 & \textbf{0.128} & \textbf{0.127} & 0.720 & 0.654 & \textbf{0.711} & 0.196 & 0.194 & 0.756 & 0.601 & 0.573 \\
\bottomrule
    \end{tabular}
    }
    \caption{Experimental results by the different prompt. \textbf{Bold} indicates the best score across the models.}
    \label{tab:overall-results-2}
\end{table*}

\begin{figure*}[t]
    \centering
    \includegraphics[width=\linewidth]{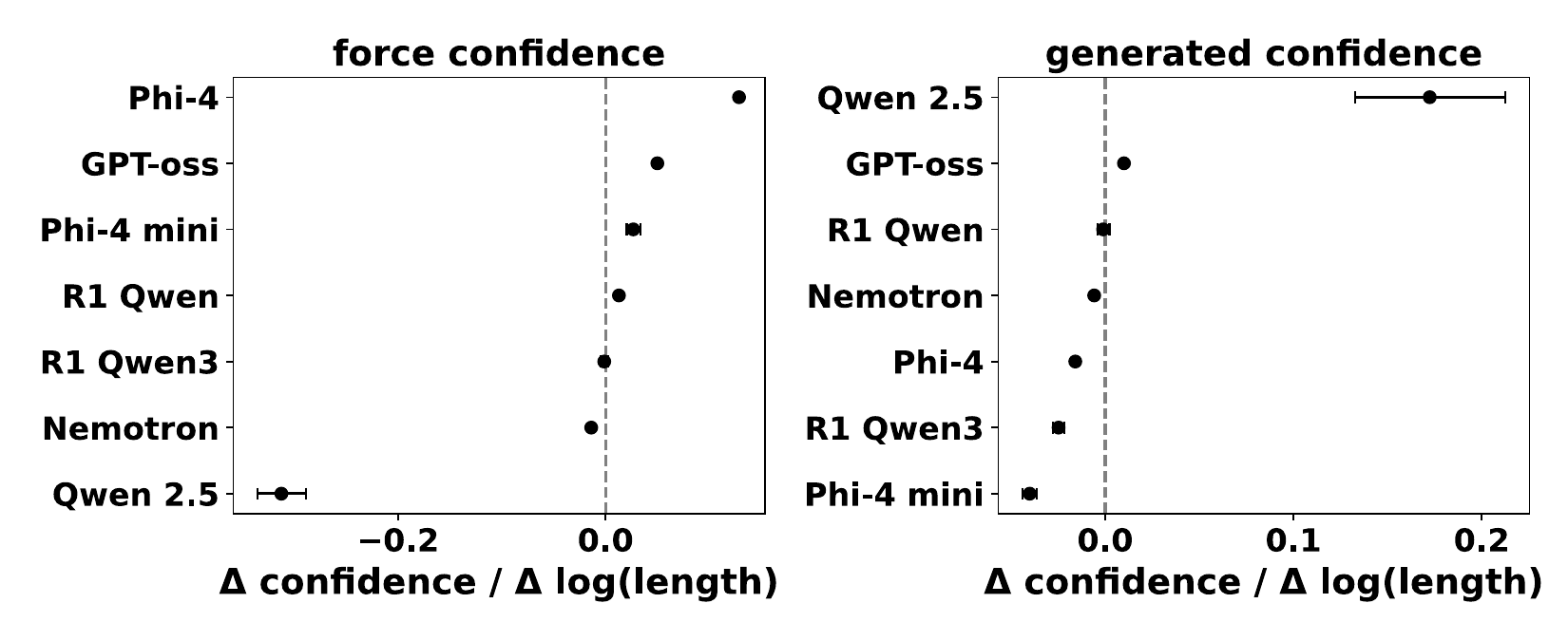}
    \caption{
    Estimated effect of reasoning length on confidence.
    Values close to zero indicate little association between changes in reasoning length and confidence.
    }
    \label{fig:paired_reasoning_confidence}
\end{figure*}

\begin{figure*}[t]
    \centering
    \includegraphics[width=\linewidth]{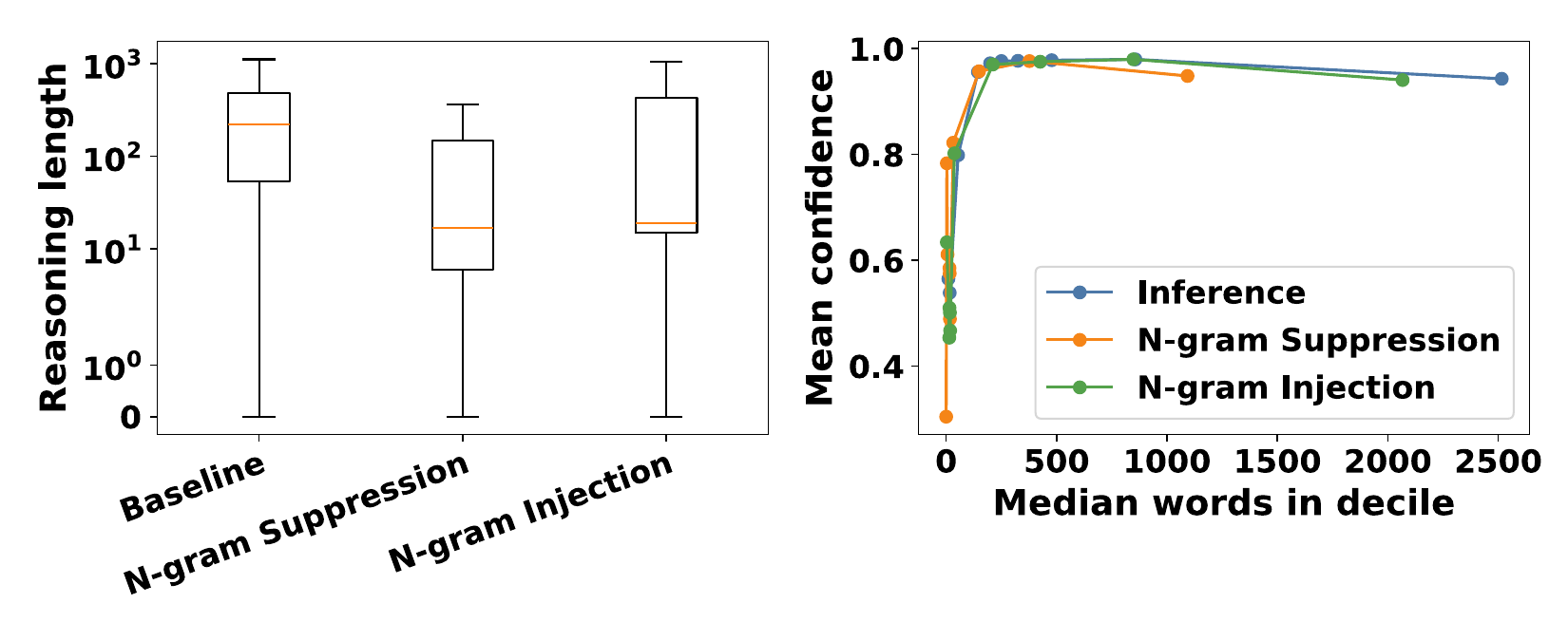}
    \caption{
    Reasoning length distributions (left) and confidence across reasoning-length deciles (right).
    Confidence remains similar despite substantial changes in reasoning length.
    }
    \label{fig:reasoning_length}
\end{figure*}

\section{\red{Effect of Reasoning Quality.}}
\red{
In order to analyze whether the effectiveness of $N$-gram Suppression stems from changes in reasoning quality rather than the suppression of confidence-related linguistic cues, we evaluate reasoning quality before and after the intervention using a Process Reward Model~\cite{zhang-etal-2025-lessons}.
Table~\ref{tab:prm-results} presents the changes in mean step scores across seven models.
We observe only marginal changes in reasoning quality, with both positive and negative differences depending on the model, suggesting that the observed changes in confidence cannot be explained solely by improvements in reasoning quality, and further supporting the role of superficial linguistic cues in shaping model confidence.
}

\section{Reasoning Length Does Not Explain Confidence Changes}
\label{app:reasoning-length-bias}
One possible alternative explanation for our findings is that the observed
confidence changes are caused by changes in reasoning length rather than by the
presence or absence of confidence-related lexical expressions.
To investigate this possibility, we first analyze whether changes in reasoning
length are consistently associated with changes in confidence.

Figure~\ref{fig:paired_reasoning_confidence} summarizes the estimated effect of
reasoning length on confidence for each model.
Specifically, the horizontal axis represents the amount of confidence change per
unit change in the logarithm of reasoning length.
If reasoning length were a major determinant of confidence, these estimates
would consistently deviate from zero across models.
Instead, most estimates remain close to zero, and their directions vary across
models.
Similar observations hold for both forced-decoding confidence and
generation-based confidence.
These results indicate that changes in reasoning length are not consistently
associated with changes in confidence.

One may argue that the lack of correlation simply reflects insufficient changes
in reasoning length.
However, Figure~\ref{fig:reasoning_length} shows that this is not the case.
Both $N$-gram Suppression and $N$-gram Injection substantially alter the length of
generated reasoning compared with the baseline.
In particular, $N$-gram Suppression produces considerably shorter reasoning
traces, while $N$-gram Injection also changes the reasoning length distribution.

Despite these substantial differences in reasoning length, the average
confidence remains nearly identical across different length bins for all three
settings.
The confidence--length curves almost completely overlap, suggesting that
responses with similar reasoning lengths exhibit similar confidence regardless
of the applied $n$-gram manipulation.

\section{Analysis on Open-ended Generation QA Tasks}
\label{app:analysis-on-open-ended-generation-qa-tasks}
Our analysis primarily focuses on multiple-choice question answering (MCQA), where confidence can be directly estimated from answer option probabilities.
To examine whether our findings generalize beyond MCQA, we extend our analysis to open-ended generation QA tasks.
Since confidence cannot be directly derived from answer choices in this setting, we evaluate model confidence using two widely adopted confidence measures: P(True)~\cite{kadavath2022languagemodelsmostlyknow}, which estimates the probability that a generated answer is correct, and sequence probability (Seq. Prob.)~\cite{lin-etal-2024-contextualized}, which measures the likelihood of the generated response.

\paragraph{P(True).}
Since open-ended generation tasks do not provide explicit answer options, confidence cannot be directly computed from option probabilities as in multiple-choice QA. Following \citet{kadavath2022languagemodelsmostlyknow}, we estimate confidence using \textit{P(True)}, which measures the probability that a model assigns to the proposition that its generated answer is correct.
Specifically, after generating an answer for a given input, we re-prompt the model with the input and its generated response, and ask whether the proposed answer is \texttt{True} or \texttt{False}. 
Our study then computes the probability assigned to the \texttt{True} option and uses it as the confidence score. 
Formally, for an input $x$ and generated answer $\hat{y}$, P(True) is defined as the probability that the model predicts $\hat{y}$ to be correct conditioned on $(x, \hat{y})$. 
A higher probability indicates that the model is more confident that its generated text output should be correct.

\paragraph{Sequence Probability (Seq-Prob).}
Following \citet{lin-etal-2024-contextualized}, we use the generation probability of an output sequence as a confidence score.
For an input $x$ and generated response $\hat{y} = [\hat{y}_1, \dots, \hat{y}_n]$, the sequence probability is defined autoregressively as the product of token conditional probabilities:

\begin{align}
    P(\hat{y}\mid x)=\prod_{i=1}^{n} P(\hat{y}_i \mid \hat{y}_{<i}, x).
\end{align}
In practice, our study converts it into log-probability to improve numerical stability:
\begin{align}
    \mathrm{SeqProb}(x,\hat{y})=\sum_{i=1}^{n}\log P(\hat{y}_i \mid \hat{y}_{<i}, x).    
\end{align}
Since longer sequences tend to have a lower likelihood, we additionally consider a length variant:
\begin{align}
\mathrm{SeqProb}_{\text{norm}}(x,\hat{y})=\frac{1}{n}\sum_{i=1}^{n}\log P(\hat{y}_i \mid \hat{y}_{<i}, x).
\end{align}
A higher sequence probability indicates that the model assigns a higher likelihood to its generated response, and thus reflects greater confidence.

\section{Generation and Forced-decoding Confidence}
\label{calcurating-confidence-by-generation-or-force-decoding}
In the generation-based method, we reuse the reasoning segment generated when solving the QA task as input, and following prior work~\citep{lin2022teaching,taubenfeld-etal-2025-confidence}, we compute confidence by prompting the model with \texttt{So, the answer is \{candidate\}. Now I will rate my confidence on a scale of 1--10. Please generate only the score. Proposed confidence: } and having it rate the reasoning segment on a 10-point scale for each candidate.
We then take the highest score as the confidence.
In our study, for reasoning models, we define the reasoning segment as the content enclosed by the \texttt{<think>} tag, while for instruction-tuned models, we define it as the content from \texttt{Let's think step by step.} proposed by \citet{kojima2022large}.
Figure~\ref{fig:calibration_curve_generation} shows the calibration curves obtained using the generation-based method, showing that the models are overconfident.
At the same time, the generation-based method tends to produce discrete values, which causes the points to scatter excessively and form a jagged curve.
Based on these results, in our study, we adopt forced decoding.

\begin{table*}[t]
    \centering
    \setlength{\tabcolsep}{1pt}
    \renewcommand{\arraystretch}{1.2}
    \rowcolors{6}{gray!10}{white}
    \resizebox{\linewidth}{!}{
    \begin{tabular}{c ccccc ccccc ccccc}
    \toprule
    \multirow{2.25}{*}{\textbf{Model}} & \multicolumn{5}{c}{\textbf{Baseline Inference}} & \multicolumn{5}{c}{\textbf{$N$-gram Suppression}} & \multicolumn{5}{c}{\textbf{$N$-gram Injection}} \\
    \cmidrule(lr){2-6}\cmidrule(lr){7-11}\cmidrule(lr){12-16}
     & \textbf{ECE $\downarrow$} & \textbf{ACE $\downarrow$} & \textbf{Brier $\downarrow$} & \textbf{Acc. $\uparrow$} & \textbf{AUROC $\uparrow$}
     & \textbf{ECE $\downarrow$} & \textbf{ACE $\downarrow$} & \textbf{Brier $\downarrow$} & \textbf{Acc. $\uparrow$} & \textbf{AUROC $\uparrow$}
     & \textbf{ECE $\downarrow$} & \textbf{ACE $\downarrow$} & \textbf{Brier $\downarrow$} & \textbf{Acc. $\uparrow$} & \textbf{AUROC $\uparrow$} \\
    \midrule
R1 Llama & 0.245 & 0.245 & \textbf{0.615} & \textbf{0.620} & 0.691 & \textbf{0.079} & \textbf{0.079} & 0.769 & 0.251 & 0.528 & 0.267 & 0.267 & 0.723 & 0.488 & \textbf{0.728} \\
Llama 3.3 & \textbf{0.121} & \textbf{0.121} & \textbf{0.685} & \textbf{0.382} & \textbf{0.733} & 0.146 & 0.146 & 0.774 & 0.282 & 0.599 & 0.148 & 0.148 & 0.802 & 0.240 & 0.522 \\
    \bottomrule
    \end{tabular}
    }
    \caption{Failure results in our experiments by Llama. Bold text indicates the best score among the models.}
    \label{tab:failure-results}
\end{table*}

\section{Sensitivity to Feature Representation in Regression Analysis}
\label{app:sensitivity-to-feature-representation}
Regarding the construction of features in \S\ref{methods}, we conducted our discussion based on frequency.
Another potential approach for feature creation involves a TF-IDF-based approach.
Demonstrating a correlation between these methods would strengthen the claim that features contributing to confidence exist regardless of the specific representation.
Figure~\ref{fig:correlation-between-different-features} shows the degree of consistency between TF-IDF-based and count-based features.
Overall, the Spearman correlation shows positive values across most models and $n$-gram sizes, confirming that the ranking of important $n$-grams remains largely consistent even when the feature construction method differs, suggesting that the regression analysis in our work captures similar trends for both frequency-based and TF-IDF-based representations, and that results do not change significantly due to feature design.
Furthermore, looking at the top-$k$ overlap using the Jaccard coefficient, particularly Jaccard@500, we observe a constant level of consistency in many settings, indicating that the majority of important feature vocabularies are shared.
While the consistency rate at Jaccard@100 decreases and settings, we attribute this to the swapping of a very small number of top feature words.

\section{Differentiation by Prompting Methods: With/Without Influential $N$-grams}
\label{app:differentiation-by-prompting-methods}
As discussed in \S\ref{causal-relationship}, we showed that manipulating specific $n$-grams through prompting methods can control confidence.
In contrast, suppression-based methods previously informed the model that producing such expressions leads to overconfidence, which explicitly conveyed that the model was overconfident.
Thus, we tested whether simply instructing the model not to produce such expressions and reducing their frequency would change the results.
Table~\ref{tab:overall-results-2} reports the results of this experiment (Suppression only $N$-gram).
Instruction-tuned models and reasoning models exhibit similar tendencies.
Based on this observation, we conducted this experiment on higher-performing reasoning models.
The results show that we can mitigate overconfidence even without explicitly stating that the model is overconfident.
Specifically, for R1 Qwen, ACE decreases from 0.334 in the Baseline to 0.323 in Suppression-only $N$-gram.
For R1 Llama, although accuracy decreases, ACE improves from 0.245 to 0.088, and we observe similar improvements for other models.

\section{Failure Cases}
\label{app:failure-cases}
Following ARR guidelines, we acknowledge that our analysis method does not generalize equally well across all model families.\footnote{C.f., H6. The results are negative. \url{https://aclrollingreview.org/reviewerguidelines}}
In addition to the models presented in the main paper, we also evaluate our method on the reasoning model DeepSeek R1 Llama (denoted as R1 Llama)~\cite{guo2025deepseek} and the instruction-tuned model Llama 3.3 70B (denoted as Llama 3.3)~\cite{grattafiori2024llama}, and observe cases where our approach is less effective. The detailed results are reported in Table~\ref{tab:failure-results}.
In particular, Llama 3.3 shows degradation across all evaluation metrics, with ACE increasing from 0.121 to 0.146 and accuracy decreasing from 0.382 to 0.282. This observation is consistent with prior studies suggesting that confidence calibration methods tend to work well for Qwen and Phi models but are less effective for Llama models~\citep{ozaki-etal-2025-understanding}. It also aligns with findings that Qwen exhibits stronger reasoning behavior than Llama~\citep{gandhi2025cognitive}, suggesting that Llama may rely on a different reasoning mechanism or reasoning style.
Despite these failure cases, our method consistently improves calibration in most models evaluated in the main paper (Table~\ref{tab:overall-results}) by suppressing $n$-grams associated with overconfidence, suggesting that while the effectiveness of our approach is model-dependent, it provides useful insights into the relationship between them.

\section{\red{Random $N$-gram Suppression}}
\red{
To analyze whether the effectiveness of $N$-gram Suppression stems from the identified confidence-related n-grams rather than simply restricting the generation of arbitrary expressions, we conduct an additional experiment in which randomly selected n-grams are suppressed instead of those identified through regression analysis. 
Table~\ref{tab:random-ngram-suppression} presents the changes in evaluation metrics relative to the baseline. We observe that random n-gram suppression consistently degrades performance across all seven models, increasing ECE, ACE, and Brier scores while decreasing accuracy and AUROC. 
In particular, accuracy drops substantially for several models, suggesting that suppressing arbitrary n-grams can disrupt the reasoning process without improving confidence calibration. 
These findings suggest that the effectiveness of our approach does not arise merely from restricting n-gram generation, but rather from selectively suppressing linguistic expressions associated with increased confidence. 
}

\begin{table}[t]
\centering
\resizebox{\linewidth}{!}{
\begin{tabular}{lrrrrr}
\toprule
Model & $\Delta$ ECE & $\Delta$ ACE & $\Delta$ Brier & $\Delta$ Acc. & $\Delta$ AUROC \\
\midrule
R1 Qwen    & +0.022 & +0.021 & +0.046 &  -2.2 & -0.084 \\
Nemotron   & +0.057 & +0.049 & +0.173 & -18.4 & -0.091 \\
R1 Qwen3   & +0.073 & +0.073 & +0.234 & -39.6 & -0.120 \\
Phi-4      & +0.003 & +0.005 & +0.382 & -46.7 & -0.132 \\
GPT-oss    & +0.031 & +0.028 & +0.119 & -12.8 & -0.067 \\
Qwen2.5    & +0.064 & +0.058 & +0.207 & -27.5 & -0.104 \\
Phi-4 mini & +0.041 & +0.036 & +0.151 & -21.3 & -0.076 \\
\bottomrule
\end{tabular}
}
\caption{Performance changes from the baseline under random n-gram suppression. Positive values indicate improvement and negative values indicate degradation for every metric.}
\label{tab:random-ngram-suppression}
\end{table}

\begin{table*}[t]
    \centering
    \renewcommand{\arraystretch}{0.9}
    \rowcolors{14}{gray!10}{white}
    \setlength{\tabcolsep}{8pt}
    \resizebox{\textwidth}{!}{
    \begin{tabular}{cc ccc ccc ccc}
    \toprule
    \multirow{3.75}{*}{\textbf{Model}} & \multirow{3.75}{*}{\textbf{$N$}} & \multicolumn{3}{c}{\textbf{Acc.}} & \multicolumn{6}{c}{\textbf{Confidence}} \\
    \cmidrule(lr){3-5} \cmidrule(lr){6-11}
    & & \multirow{2.25}{*}{\textbf{Acc.}} & \multirow{2.25}{*}{\shortstack{\textbf{ROC} \\ \textbf{AUC}}} & \multirow{2.25}{*}{\shortstack{\textbf{Log} \\ \textbf{Loss}}} & \multicolumn{3}{c}{\textbf{Generation}} & \multicolumn{3}{c}{\textbf{Forced-Decoding}} \\
    \cmidrule(lr){6-8}\cmidrule(lr){9-11}
     &  & & & & \textbf{$\lambda$} & \textbf{MAE} & \textbf{MRE} & \textbf{$\lambda$} & \textbf{MAE} & \textbf{MRE}  \\
    \midrule
 & 1 & 0.6150 & 0.5000 & 0.7541 & 0.0000 & 0.1913 & 0.4531 & 0.0000 & 0.0880 & 0.1336 \\
 & 2 & 0.6150 & 0.5000 & 0.7540 & 0.0000 & 0.1871 & 0.4396 & 0.0046 & 0.1900 & 0.3991 \\
 & 3 & 0.6150 & 0.5000 & 0.7540 & 0.0000 & 0.2191 & 0.5117 & 0.0029 & 0.1866 & 0.3922 \\
 & 4 & 0.6150 & 0.5000 & 0.7540 & 0.0000 & 0.2467 & 0.5733 & 0.0018 & 0.1900 & 0.3991 \\
 & 5 & 0.6150 & 0.5000 & 0.7540 & 0.0000 & 0.2606 & 0.6010 & 0.0016 & 0.1900 & 0.3991 \\
 & 6 & 0.6150 & 0.5000 & 0.7540 & 0.0000 & 0.2814 & 0.6493 & 0.0008 & 0.1900 & 0.3991 \\
 & 7 & 0.6150 & 0.5000 & 0.7540 & 0.0000 & 0.2949 & 0.6809 & 0.0003 & 0.1900 & 0.3991 \\
\multirow{-8}{*}{R1 Llama} & 8 & 0.6150 & 0.5000 & 0.7540 & 0.0000 & 0.3016 & 0.6966 & 0.0003 & 0.1900 & 0.3991 \\
\cmidrule(lr){2-11}
 & 1 & 0.6175 & 0.5000 & 0.7433 & 0.0001 & 0.2505 & 0.5910 & 0.0000 & 0.0474 & 0.0652 \\
 & 2 & 0.6175 & 0.5000 & 0.7433 & 0.0000 & 0.2410 & 0.5683 & 0.0000 & 0.0452 & 0.0619 \\
 & 3 & 0.6395 & 0.6202 & 0.6438 & 0.0000 & 0.2443 & 0.5754 & 0.0000 & 0.0448 & 0.0611 \\
 & 4 & 0.6175 & 0.5000 & 0.7433 & 0.0000 & 0.2533 & 0.5956 & 0.0000 & 0.0486 & 0.0666 \\
 & 5 & 0.6175 & 0.5000 & 0.7433 & 0.0000 & 0.2549 & 0.5988 & 0.0000 & 0.0481 & 0.0653 \\
 & 6 & 0.6175 & 0.5000 & 0.7433 & 0.0000 & 0.2670 & 0.6271 & 0.0000 & 0.0508 & 0.0691 \\
 & 7 & 0.6175 & 0.5000 & 0.7433 & 0.0000 & 0.2643 & 0.6215 & 0.0000 & 0.0507 & 0.0690 \\
\multirow{-8}{*}{R1 Qwen} & 8 & 0.6186 & 0.5163 & 0.6613 & 0.0000 & 0.2720 & 0.6397 & 0.0000 & 0.0521 & 0.0706 \\
\cmidrule(lr){2-11}
 & 1 & 0.6246 & 0.6686 & 0.6500 & 0.0001 & 0.2744 & 0.6442 & 0.0001 & 0.0744 & 0.0997 \\
 & 2 & 0.5476 & 0.5719 & 0.6853 & 0.0000 & 0.2717 & 0.6370 & 0.0000 & 0.0739 & 0.0990 \\
 & 3 & 0.5236 & 0.5403 & 0.6889 & 0.0001 & 0.2863 & 0.6699 & 0.0000 & 0.0748 & 0.1002 \\
 & 4 & 0.5144 & 0.5000 & 0.6927 & 0.0001 & 0.2907 & 0.6801 & 0.0000 & 0.0755 & 0.1014 \\
 & 5 & 0.5185 & 0.5237 & 0.6892 & 0.0001 & 0.2941 & 0.6884 & 0.0000 & 0.0758 & 0.1016 \\
 & 6 & 0.4856 & 0.5000 & 0.7050 & 0.0001 & 0.2946 & 0.6897 & 0.0000 & 0.0749 & 0.1002 \\
 & 7 & 0.4856 & 0.5000 & 0.7050 & 0.0001 & 0.2947 & 0.6901 & 0.0000 & 0.0760 & 0.1020 \\
\multirow{-8}{*}{Nemotron} & 8 & 0.4856 & 0.5000 & 0.7050 & 0.0001 & 0.2947 & 0.6901 & 0.0000 & 0.0759 & 0.1018 \\
\cmidrule(lr){2-11}
 & 1 & 0.7438 & 0.5000 & 0.5690 & 0.0001 & 0.1979 & 0.4326 & 0.0000 & 0.0419 & 0.0503 \\
 & 2 & 0.7438 & 0.5000 & 0.5690 & 0.0000 & 0.1948 & 0.4263 & 0.0000 & 0.0394 & 0.0472 \\
 & 3 & 0.7438 & 0.6742 & 0.5456 & 0.0000 & 0.1953 & 0.4274 & 0.0000 & 0.0378 & 0.0451 \\
 & 4 & 0.7438 & 0.5000 & 0.5690 & 0.0000 & 0.1954 & 0.4268 & 0.0000 & 0.0374 & 0.0446 \\
 & 5 & 0.7438 & 0.5000 & 0.5690 & 0.0000 & 0.1961 & 0.4276 & 0.0000 & 0.0374 & 0.0446 \\
 & 6 & 0.7438 & 0.5000 & 0.5690 & 0.0000 & 0.1992 & 0.4335 & 0.0000 & 0.0387 & 0.0464 \\
 & 7 & 0.7438 & 0.5000 & 0.5690 & 0.0000 & 0.2039 & 0.4433 & 0.0000 & 0.0396 & 0.0476 \\
\multirow{-8}{*}{R1 Qwen3} & 8 & 0.7438 & 0.5000 & 0.5690 & 0.0000 & 0.2043 & 0.4436 & 0.0000 & 0.0395 & 0.0474 \\
\cmidrule(lr){2-11}
 & 1 & 0.7930 & 0.5000 & 0.5580 & 0.0000 & 0.1669 & 0.3463 & 0.0000 & 0.0057 & 0.0062 \\
 & 2 & 0.7930 & 0.5000 & 0.5580 & 0.0000 & 0.1595 & 0.3323 & 0.0000 & 0.0056 & 0.0061 \\
 & 3 & 0.7930 & 0.5000 & 0.5580 & 0.0000 & 0.1650 & 0.3452 & 0.0000 & 0.0057 & 0.0062 \\
 & 4 & 0.7930 & 0.5000 & 0.5580 & 0.0000 & 0.1712 & 0.3587 & 0.0000 & 0.0057 & 0.0062 \\
 & 5 & 0.7930 & 0.5000 & 0.5580 & 0.0000 & 0.1721 & 0.3585 & 0.0000 & 0.0058 & 0.0063 \\
 & 6 & 0.7930 & 0.5000 & 0.5580 & 0.0000 & 0.1835 & 0.3834 & 0.0000 & 0.0060 & 0.0065 \\
 & 7 & 0.7930 & 0.5000 & 0.5580 & 0.0000 & 0.1895 & 0.3951 & 0.0000 & 0.0060 & 0.0065 \\
\multirow{-8}{*}{Phi-4} & 8 & 0.7930 & 0.5000 & 0.5580 & 0.0000 & 0.2012 & 0.4204 & 0.0000 & 0.0063 & 0.0068 \\
\cmidrule(lr){2-11}
 & 1 & 0.8035 & 0.7668 & 0.4491 & 0.0001 & 0.2852 & 0.7214 & 0.0000 & 0.0209 & 0.0234 \\
 & 2 & 0.7775 & 0.5000 & 0.5305 & 0.0000 & 0.2716 & 0.6888 & 0.0000 & 0.0360 & 0.0484 \\
 & 3 & 0.7775 & 0.5000 & 0.5305 & 0.0000 & 0.2821 & 0.7192 & 0.0000 & 0.0334 & 0.0468 \\
 & 4 & 0.7775 & 0.5000 & 0.5305 & 0.0000 & 0.2984 & 0.7624 & 0.0000 & 0.0318 & 0.0455 \\
 & 5 & 0.7775 & 0.5000 & 0.5305 & 0.0000 & 0.3269 & 0.8367 & 0.0000 & 0.0308 & 0.0450 \\
 & 6 & 0.7786 & 0.5649 & 0.5228 & 0.0002 & 0.3557 & 0.9108 & 0.0000 & 0.0311 & 0.0455 \\
 & 7 & 0.7775 & 0.5000 & 0.5305 & 0.0001 & 0.3565 & 0.9130 & 0.0000 & 0.0318 & 0.0464 \\
\multirow{-8}{*}{GPT-oss} & 8 & 0.7775 & 0.5000 & 0.5305 & 0.0000 & 0.3466 & 0.8874 & 0.0000 & 0.0319 & 0.0465 \\
\bottomrule
    \end{tabular}
    }
    \caption{Evaluation results of the regression analysis. We use accuracy, ROC-AUC, and log loss to evaluate logistic regression, and MAE and MRE to evaluate Lasso regression. $\lambda$ denotes the regularization coefficient of Lasso.}
    \label{tab:lasso-regression-evaluation-1}
\end{table*}

\begin{table*}[t]
    \centering
    \rowcolors{14}{gray!10}{white}
    \setlength{\tabcolsep}{8pt}
    \resizebox{\textwidth}{!}{
    \begin{tabular}{cc ccc ccc ccc}
    \toprule
    \multirow{3.75}{*}{\textbf{Model}} & \multirow{3.75}{*}{\textbf{$N$}} & \multicolumn{3}{c}{\textbf{Acc.}} & \multicolumn{6}{c}{\textbf{Confidence}} \\
    \cmidrule(lr){3-5} \cmidrule(lr){6-11}
    & & \multirow{2.25}{*}{\textbf{Acc.}} & \multirow{2.25}{*}{\shortstack{\textbf{ROC} \\ \textbf{AUC}}} & \multirow{2.25}{*}{\shortstack{\textbf{Log} \\ \textbf{Loss}}} & \multicolumn{3}{c}{\textbf{Generation}} & \multicolumn{3}{c}{\textbf{Forced-Decoding}} \\
    \cmidrule(lr){6-8}\cmidrule(lr){9-11}
     &  & & & & \textbf{$\lambda$} & \textbf{MAE} & \textbf{MRE} & \textbf{$\lambda$} & \textbf{MAE} & \textbf{MRE}  \\
    \midrule
 & 1 & 0.7604 & 0.6407 & 0.5419 & 0.0001 & 0.2514 & 0.5519 & 0.0006 & 0.1434 & 0.2723 \\
 & 2 & 0.7656 & 0.6509 & 0.5346 & 0.0001 & 0.2514 & 0.5531 & 0.0000 & 0.1202 & 0.2259 \\
 & 3 & 0.7642 & 0.6387 & 0.5375 & 0.0000 & 0.2509 & 0.5528 & 0.0000 & 0.1225 & 0.2307 \\
 & 4 & 0.7893 & 0.7439 & 0.4852 & 0.0001 & 0.2564 & 0.5672 & 0.0000 & 0.1307 & 0.2464 \\
 & 5 & 0.7753 & 0.6847 & 0.5108 & 0.0000 & 0.2545 & 0.5627 & 0.0000 & 0.1562 & 0.2950 \\
 & 6 & 0.7620 & 0.6277 & 0.5187 & 0.0000 & 0.2537 & 0.5610 & 0.0000 & 0.1744 & 0.3302 \\
 & 7 & 0.7475 & 0.5671 & 0.5505 & 0.0000 & 0.2567 & 0.5681 & 0.0000 & 0.1867 & 0.3541 \\
\multirow{-8}{*}{Llama3.3} & 8 & 0.7373 & 0.5297 & 0.5703 & 0.0000 & 0.2595 & 0.5742 & 0.0000 & 0.1933 & 0.3669 \\
\cmidrule(lr){2-11}
 & 1 & 0.7754 & 0.7391 & 0.5328 & 0.0001 & 0.1936 & 0.4403 & 0.0001 & 0.0851 & 0.1863 \\
 & 2 & 0.7507 & 0.7238 & 0.5508 & 0.0001 & 0.1968 & 0.4485 & 0.0000 & 0.0936 & 0.1976 \\
 & 3 & 0.6961 & 0.6615 & 0.6005 & 0.0001 & 0.1986 & 0.4513 & 0.0000 & 0.1225 & 0.2532 \\
 & 4 & 0.6750 & 0.6238 & 0.6190 & 0.0001 & 0.2020 & 0.4589 & 0.0000 & 0.1338 & 0.2752 \\
 & 5 & 0.6557 & 0.6405 & 0.6100 & 0.0001 & 0.2034 & 0.4617 & 0.0000 & 0.1584 & 0.3229 \\
 & 6 & 0.6217 & 0.5196 & 0.6609 & 0.0005 & 0.2071 & 0.4708 & 0.0000 & 0.1757 & 0.3576 \\
 & 7 & 0.6175 & 0.5000 & 0.6746 & 0.0000 & 0.1999 & 0.4528 & 0.0000 & 0.1859 & 0.3786 \\
\multirow{-8}{*}{Qwen2.5} & 8 & 0.6175 & 0.5000 & 0.6746 & 0.0000 & 0.2030 & 0.4600 & 0.0000 & 0.1914 & 0.3900 \\
\cmidrule(lr){2-11}
 & 1 & 0.6598 & 0.5000 & 0.6414 & 0.0001 & 0.2024 & 0.4427 & 0.0000 & 0.0554 & 0.0851 \\
 & 2 & 0.6598 & 0.5000 & 0.6412 & 0.0000 & 0.1977 & 0.4324 & 0.0000 & 0.0490 & 0.0739 \\
 & 3 & 0.6598 & 0.5000 & 0.6412 & 0.0000 & 0.2014 & 0.4391 & 0.0000 & 0.0481 & 0.0724 \\
 & 4 & 0.6598 & 0.5000 & 0.6412 & 0.0000 & 0.2090 & 0.4526 & 0.0000 & 0.0477 & 0.0716 \\
 & 5 & 0.6598 & 0.5000 & 0.6934 & 0.0000 & 0.2169 & 0.4681 & 0.0000 & 0.0499 & 0.0753 \\
 & 6 & 0.6598 & 0.5000 & 0.6413 & 0.0007 & 0.2400 & 0.5186 & 0.0000 & 0.0515 & 0.0778 \\
 & 7 & 0.6598 & 0.5000 & 0.6413 & 0.0005 & 0.2400 & 0.5186 & 0.0000 & 0.0555 & 0.0845 \\
\multirow{-8}{*}{Phi-4-mini} & 8 & 0.6598 & 0.5000 & 0.6413 & 0.0005 & 0.2400 & 0.5186 & 0.0000 & 0.0569 & 0.0869 \\
\bottomrule
    \end{tabular}
    }
    \caption{Evaluation results of the regression analysis. We use accuracy, ROC-AUC, and log loss to evaluate logistic regression, and MAE and MRE to evaluate Lasso regression. $\lambda$ denotes the regularization coefficient of Lasso.}
    \label{tab:lasso-regression-evaluation-2}
\end{table*}

\begin{table*}[t]
    \centering
    \renewcommand{\arraystretch}{1.2}
    \setlength{\tabcolsep}{1pt}
    \rowcolors{4}{gray!10}{white}
    \resizebox{\textwidth}{!}{
    \begin{tabular}{crrr crrr crrr crrr}
    \toprule
    \textbf{1-gram} &  \textbf{Conf.} & \textbf{Acc.} & \textbf{Freq.} &
    \textbf{2-gram} &  \textbf{Conf.} & \textbf{Acc.} & \textbf{Freq.} &
    \textbf{3-gram} &  \textbf{Conf.} & \textbf{Acc.} & \textbf{Freq.} &
    \textbf{4-gram} &  \textbf{Conf.} & \textbf{Acc.} & \textbf{Freq.} \\
\cmidrule(lr){1-16}
\rowcolor{blue!10}
\multicolumn{16}{c}{\textbf{R1 Qwen3}} \\
\cmidrule(lr){1-16}
directly & 1.32 & 0.00$^\dagger$ & 96965 & decision select & 3.23 & 0.00$^\dagger$ & 1167 & final decision select & 2.55 & 0.15$^\dagger$ & 1164 & choice final decision select & 1.80 & 0.00$^\dagger$ & 158 \\
yes & 1.22 & 0.00$^\dagger$ & 8209 & add explanation & 2.79 & 0.00$^\dagger$ & 2258 & user asking answer & 1.58 & 0.15$^\dagger$ & 2044 & perhaps need output thought & 1.60 & 0.00$^\dagger$ & 351 \\
output & 0.94 & 0.00$^\dagger$ & 156404 & asking answer & 2.11 & 0.00$^\dagger$ & 2046 & says add explanation & 1.57 & 0.15$^\dagger$ & 723 & asking answer question selecting & 1.54 & 0.00$^\dagger$ & 1666 \\
context & 0.90 & 0.00$^\dagger$ & 153054 & let verify & 1.80 & 0.00$^\dagger$ & 2916 & perhaps need output & 1.55 & 0.15$^\dagger$ & 2010 & guidelines include having thought & 1.42 & 0.00$^\dagger$ & 1300 \\
probably & 0.90 & 0.00$^\dagger$ & 27043 & need infer & 1.57 & 0.00$^\dagger$ & 7288 & exactly option question & 1.45 & 0.15$^\dagger$ & 2660 & explanation output required format & 1.42 & 0.00$^\dagger$ & 738 \\
say & 0.71 & 0.00$^\dagger$ & 280431 & match exactly & 1.48 & 0.00$^\dagger$ & 7129 & incomplete need infer & 1.40 & 0.15$^\dagger$ & 1048 & select option correctly describes & 1.34 & 0.00$^\dagger$ & 821 \\
number & 0.68 & 0.00$^\dagger$ & 15250 & correctly describes & 1.40 & 0.00$^\dagger$ & 3049 & option option talks & 1.37 & 0.15$^\dagger$ & 820 & concise detailing necessary steps & 1.34 & 0.00$^\dagger$ & 476 \\
knife & 0.65 & 0.00$^\dagger$ & 9491 & main sections & 1.38 & 0.00$^\dagger$ & 11498 & option correctly describes & 1.37 & 0.15$^\dagger$ & 2757 & output final answer think & 1.32 & 0.00$^\dagger$ & 915 \\
dog & 0.40 & 0.00$^\dagger$ & 15258 & option correctly & 1.29 & 0.00$^\dagger$ & 5649 & user said add & 1.29 & 0.15$^\dagger$ & 485 & thought solution sections output & 1.31 & 0.00$^\dagger$ & 584 \\
reason & 0.39 & 0.00$^\dagger$ & 40114 & question guidelines & 1.23 & 0.00$^\dagger$ & 15087 & include having thought & 1.25 & 0.15$^\dagger$ & 1337 & analyze option option talks & 1.31 & 0.00$^\dagger$ & 716 \\
select & -4.98 & 0.00$^\dagger$ & 79260 & think select & -6.20 & 0.00$^\dagger$ & 3314 & \textcolor{blue}{\textbf{water soaked cotton}} & \textcolor{blue}{\textbf{-5.05}} & \textcolor{blue}{\textbf{0.15}}$^\dagger$ & \textcolor{blue}{\textbf{46}} & think make choice let & -6.84 & 0.00$^\dagger$ & 199 \\
choose & -2.34 & 0.00$^\dagger$ & 58403 & answer whichever & -4.18 & 0.00$^\dagger$ & 698 & \textcolor{blue}{\textbf{doesn specify season}} & \textcolor{blue}{\textbf{-4.82}} & \textcolor{blue}{\textbf{0.15}}$^\dagger$ & \textcolor{blue}{\textbf{17}} & advice based personal experience & -5.82 & 0.00$^\dagger$ & 2 \\
perhaps & -1.45 & 0.00$^\dagger$ & 195391 & think ll & -3.34 & 0.00$^\dagger$ & 4795 & \textcolor{blue}{\textbf{assistant helps people}} & \textcolor{blue}{\textbf{-4.69}} & \textcolor{blue}{\textbf{0.15}}$^\dagger$ & \textcolor{blue}{\textbf{2}} & ai assistant helps people & -5.06 & 0.00$^\dagger$ & 2 \\
think & -1.08 & 0.00$^\dagger$ & 172014 & need accept & -2.71 & 0.00$^\dagger$ & 1870 & \textcolor{blue}{\textbf{basketball gym man}} & \textcolor{blue}{\textbf{-4.33}} & \textcolor{blue}{\textbf{0.15}}$^\dagger$ & \textcolor{blue}{\textbf{28}} & using mixer puts ingredients & -5.02 & 0.00$^\dagger$ & 33 \\
consider & -0.49 & 0.00$^\dagger$ & 42322 & blue line & -2.57 & 0.00$^\dagger$ & 103 & \textcolor{blue}{\textbf{let write response}} & \textcolor{blue}{\textbf{-4.31}} & \textcolor{blue}{\textbf{0.15}}$^\dagger$ & \textcolor{blue}{\textbf{99}} & correct option output specified & -4.61 & 0.00$^\dagger$ & 7 \\
section & -0.44 & 0.00$^\dagger$ & 167622 & french toast & -2.16 & 0.00$^\dagger$ & 80 & \textcolor{blue}{\textbf{think wasting time}} & \textcolor{blue}{\textbf{-4.25}} & \textcolor{blue}{\textbf{0.15}}$^\dagger$ & \textcolor{blue}{\textbf{554}} & day night away home & -4.10 & 0.00$^\dagger$ & 3 \\
option & -0.42 & 0.00$^\dagger$ & 867604 & section ll & -2.13 & 0.00$^\dagger$ & 2918 & \textcolor{blue}{\textbf{correct option add}} & \textcolor{blue}{\textbf{-4.17}} & \textcolor{blue}{\textbf{0.15}}$^\dagger$ & \textcolor{blue}{\textbf{11}} & think wasting time let & -3.95 & 0.00$^\dagger$ & 235 \\
playing & -0.37 & 0.00$^\dagger$ & 21347 & perhaps correct & -2.01 & 0.00$^\dagger$ & 8948 & \textcolor{blue}{\textbf{make choice let}} & \textcolor{blue}{\textbf{-4.04}} & \textcolor{blue}{\textbf{0.15}}$^\dagger$ & \textcolor{blue}{\textbf{280}} & option add extra text & -3.86 & 0.00$^\dagger$ & 2 \\
ha & -0.30 & 0.00$^\dagger$ & 74486 & make decision & -1.99 & 0.00$^\dagger$ & 2022 & \textcolor{blue}{\textbf{thought section ll}} & \textcolor{blue}{\textbf{-3.48}} & \textcolor{blue}{\textbf{0.15}}$^\dagger$ & \textcolor{blue}{\textbf{1014}} & think wasting time need & -3.79 & 0.00$^\dagger$ & 219 \\
advice & -0.16 & 0.00$^\dagger$ & 37850 & honey bee & -1.99 & 0.00$^\dagger$ & 78 & \textcolor{blue}{\textbf{output don know}} & \textcolor{blue}{\textbf{-3.40}} & \textcolor{blue}{\textbf{0.15}}$^\dagger$ & \textcolor{blue}{\textbf{11}} & blue line pushes puck & -3.39 & 0.00$^\dagger$ & 18 \\
\cmidrule(lr){1-16}
\rowcolor{blue!10}
\multicolumn{16}{c}{\textbf{GPT-oss}} \\
\cmidrule(lr){1-16}
tag & 3.82 & 0.19$^\dagger$ & 440 & assistantfinalfinal answer & 4.21 & 0.00$^\dagger$ & 10094 & answer assistantfinalfinal answer & 2.20 & 0.00$^\dagger$ & 7359 & need choose plausible continuation & 1.39 & 0.00$^\dagger$ & 85 \\
assistantfinal & 2.66 & 0.19$^\dagger$ & 8332 & answerassistantfinalfinal answer & 3.78 & 0.00$^\dagger$ & 404 & let sassistantfinalfinal answer & 2.08 & 0.00$^\dagger$ & 491 & need choose correct continuation & 1.36 & 0.00$^\dagger$ & 172 \\
sassistantfinal & 1.81 & 0.19$^\dagger$ & 431 & let parse & 3.38 & 0.00$^\dagger$ & 354 & need interpret question & 1.89 & 0.00$^\dagger$ & 369 & let output assistantfinalfinal answer & 1.32 & 0.00$^\dagger$ & 124 \\
reasoning & 0.74 & 0.19$^\dagger$ & 860 & soassistantfinalfinal answer & 3.27 & 0.00$^\dagger$ & 80 & choose assistantfinalfinal answer & 1.74 & 0.00$^\dagger$ & 119 & let answer assistantfinalfinal answer & 1.31 & 0.00$^\dagger$ & 97 \\
thinking & 0.73 & 0.19$^\dagger$ & 514 & let sassistantfinalfinal & 3.19 & 0.00$^\dagger$ & 492 & final assistantfinalfinal answer & 1.67 & 0.00$^\dagger$ & 468 & final answer assistantfinalfinal answer & 1.26 & 0.00$^\dagger$ & 5286 \\
thought & 0.71 & 0.19$^\dagger$ & 1412 & assistantfinalassistantfinal answer & 2.91 & 0.00$^\dagger$ & 350 & output assistantfinalfinal answer & 1.66 & 0.00$^\dagger$ & 326 & let produce assistantfinalfinal answer & 1.24 & 0.00$^\dagger$ & 289 \\
passage & 0.69 & 0.19$^\dagger$ & 12492 & need interpret & 2.62 & 0.00$^\dagger$ & 982 & let assistantfinalfinal answer & 1.59 & 0.00$^\dagger$ & 206 & answer let assistantfinalfinal answer & 1.20 & 0.00$^\dagger$ & 156 \\
let & 0.62 & 0.19$^\dagger$ & 39686 & assistantfinal answer & 2.49 & 0.00$^\dagger$ & 3294 & assistantfinal answerfinal answer & 1.52 & 0.00$^\dagger$ & 882 & answer let sassistantfinalfinal answer & 1.18 & 0.00$^\dagger$ & 400 \\
phrase & 0.54 & 0.19$^\dagger$ & 6564 & need infer & 2.47 & 0.00$^\dagger$ & 587 & output assistantfinal answer & 1.45 & 0.00$^\dagger$ & 205 & answer answer assistantfinalfinal answer & 1.16 & 0.00$^\dagger$ & 662 \\
\textcolor{red}{\underline{maybe}} & \textcolor{red}{\underline{0.52}} & \textcolor{red}{\underline{-1.84}} & \textcolor{red}{\underline{47610}} & answerfinal answer & 2.44 & 0.00$^\dagger$ & 1097 & accordingly assistantfinalfinal answer & 1.43 & 0.00$^\dagger$ & 487 & answer let sassistantfinal answer & 1.15 & 0.00$^\dagger$ & 136 \\
think & -6.86 & -0.59 & 17557 & density mass & -7.39 & 0.00$^\dagger$ & 5 & python assistantanalysis container & -8.39 & 0.00$^\dagger$ & 2 & policy says user requests & -12.89 & 0.00$^\dagger$ & 2 \\
\textcolor{blue}{\textbf{wethink}} & \textcolor{blue}{\textbf{-6.28}} & \textcolor{blue}{\textbf{0.19}}$^\dagger$ & \textcolor{blue}{\textbf{177}} & lines country & -6.31 & 0.00$^\dagger$ & 4 & density mass volume & -6.74 & 0.00$^\dagger$ & 2 & gives start starts ahead & -9.97 & 0.00$^\dagger$ & 4 \\
\textcolor{blue}{\textbf{thethink}} & \textcolor{blue}{\textbf{-6.05}} & \textcolor{blue}{\textbf{0.19}}$^\dagger$ & \textcolor{blue}{\textbf{88}} & outstanding balance & -5.22 & 0.00$^\dagger$ & 18 & correct answer thec & -5.57 & 0.00$^\dagger$ & 2 & listed did misinterpret maybe & -8.54 & 0.00$^\dagger$ & 2 \\
\textcolor{blue}{\textbf{installment}} & \textcolor{blue}{\textbf{-1.71}} & \textcolor{blue}{\textbf{0.19}}$^\dagger$ & \textcolor{blue}{\textbf{72}} & married couples & -4.93 & 0.00$^\dagger$ & 17 & blue colors blue & -3.07 & 0.00$^\dagger$ & 3 & answer correct answer option & -7.36 & 0.00$^\dagger$ & 10 \\
\textcolor{blue}{\textbf{user}} & \textcolor{blue}{\textbf{-1.06}} & \textcolor{blue}{\textbf{0.19}}$^\dagger$ & \textcolor{blue}{\textbf{2074}} & partitions partitions & -4.51 & 0.00$^\dagger$ & 17 & assistantfinal answer final & -0.96 & 0.00$^\dagger$ & 467 & compute need need need & -3.28 & 0.00$^\dagger$ & 3 \\
\textcolor{blue}{\textbf{thewe}} & \textcolor{blue}{\textbf{-0.97}} & \textcolor{blue}{\textbf{0.19}}$^\dagger$ & \textcolor{blue}{\textbf{2935}} & sorry help & -4.16 & 0.00$^\dagger$ & 6 & final answer assistantfinalfinal & -0.74 & 0.00$^\dagger$ & 5287 & hope help people understand & -3.18 & 0.00$^\dagger$ & 2 \\
\textcolor{blue}{\textbf{density}} & \textcolor{blue}{\textbf{-0.95}} & \textcolor{blue}{\textbf{0.19}}$^\dagger$ & \textcolor{blue}{\textbf{141}} & gives start & -3.07 & 0.00$^\dagger$ & 30 & finalfinal answer assistantfinalfinal & -0.66 & 0.00$^\dagger$ & 156 & step evaluate option option & -2.77 & 0.00$^\dagger$ & 2 \\
\textcolor{blue}{\textbf{partition}} & \textcolor{blue}{\textbf{-0.80}} & \textcolor{blue}{\textbf{0.19}}$^\dagger$ & \textcolor{blue}{\textbf{176}} & produce assistantfinalfinal & -1.59 & 0.00$^\dagger$ & 500 &  &  &  &  & step think step step & -2.58 & 0.00$^\dagger$ & 3 \\
\textcolor{blue}{\textbf{assistantfinalfinal}} & \textcolor{blue}{\textbf{-0.54}} & \textcolor{blue}{\textbf{0.19}}$^\dagger$ & \textcolor{blue}{\textbf{10096}} & density density & -1.57 & 0.00$^\dagger$ & 6 &  &  &  &  & integer integer integer integer & -2.13 & 0.00$^\dagger$ & 13 \\
\textcolor{blue}{\textbf{wewe}} & \textcolor{blue}{\textbf{-0.52}} & \textcolor{blue}{\textbf{0.19}}$^\dagger$ & \textcolor{blue}{\textbf{1018}} & people married & -1.20 & 0.00$^\dagger$ & 26 &  &  &  &  & matches final answer let & -1.45 & 0.00$^\dagger$ & 27 \\
    \bottomrule
    \end{tabular}
    }
    \caption{
    Results of models other than those in Table~\ref{tab:lasso-result-all-1}.
    }
    \label{tab:lasso-result-all-3}
\end{table*}

\section{Prompts}
\label{detailed-prompts}
We list the prompts used in the experiments below.
\onecolumn

\begin{tcolorbox}[title=Initial inference (Baseline), boxrule=1pt]
\small
You must answer this question by selecting exactly ONE of the options A, B, C, or D. \\

DO NOT: \\
- output anything except the required format, \\
- include LaTeX commands (bold, emph, etc.), \\
- change option letters, \\
- add any explanations or additional texts outside the <think> tags. \\

QUESTION: \\
\{question\} \\

OPTIONS: \\
A. {options[0]} \\ 
B. {options[1]} \\
C. {options[2]} \\
D. {options[3]} \\
E. {options[4]}  \textbf{(In the case of MathQA, the options go up to E as it has five choices.)} \\

REQUIRED OUTPUT FORMAT (must match EXACTLY): \\
Final Answer: \texttt{<OPTION>} \\

Replace \texttt{<OPTION>} with A, B, C, or D, E ONLY. \\

\end{tcolorbox}

\begin{tcolorbox}[title=Re-generation ($N$-gram Suppression), boxrule=1pt]
\small
        "IMPORTANT CONSTRAINT:" \\
        "In your reasoning process, you tend to produce overconfident answers by using certain words." \\
        "To mitigate this, please avoid using the following words or phrases in your reasoning:" \\ 
        "\{banned\_words\}." \\
        "If these words would normally appear, please rephrase your reasoning to avoid them completely." \\
\end{tcolorbox}

\begin{tcolorbox}[title=Re-generation ($N$-gram Injection), boxrule=1pt]
\small
        "IMPORTANT CONSTRAINT: \\ "
        "In your reasoning process, you tend to produce underconfident answers." \\
        "To counteract this, you should actively use confident and decisive language." \\
        "In particular, incorporate the following words or phrases naturally in your reasoning:" \\
        "\{encouraged\_words\}." \\
        "Use them appropriately to express certainty, commitment, and clear conclusions." \\
\end{tcolorbox}

\begin{tcolorbox}[title=Re-generation (Suppression only $N$-gram), boxrule=1pt]
\small
        "IMPORTANT CONSTRAINT: " \\
        "Please avoid using the following words or phrases in your reasoning. " \\
        "\{banned\_words\}. " \\
        "If these words would normally appear, please rephrase your reasoning to avoid them completely." \\
\end{tcolorbox}

\end{document}